\documentclass[10pt]{article}
\PassOptionsToPackage{table}{xcolor}
\usepackage[preprint]{tmlr}
\usepackage[utf8]{inputenc}
\usepackage[T1]{fontenc}
\usepackage{amsmath,amssymb,amsfonts}
\usepackage{booktabs}
\usepackage{graphicx}
\usepackage[table]{xcolor}
\usepackage{array}
\usepackage{flafter}
\usepackage{float}
\usepackage{wrapfig}
\usepackage[section]{placeins}
\usepackage{hyperref}
\usepackage{url}
\usepackage{microtype}
\hypersetup{hidelinks}
\definecolor{llagreen}{HTML}{1B7837}
\definecolor{llared}{HTML}{B2182B}
\definecolor{llahl}{HTML}{E3EEF9}
\newcommand{\lla}{LLA}

\newcommand{\TableTier}{%
\begin{table}[H]
  \centering\footnotesize
  \caption{\textbf{GSM8K accuracy-compression Pareto for Ouro-1.4B.} Six cold-start LLA codecs are trained with the same recipe and evaluated with the cached reconstruction decoder. Each multiplier is the exact KV-cache compression ratio $\rho$ (Eq.~\eqref{eq:ratio}). The plateau from $4.0\times$ to $2.67\times$ breaks sharply at $2.0\times$.}
  \label{tab:tier4}
  \begin{tabular}{lccc}
    \toprule
    Model & $(r_k,r_v)$ & GSM8K strict $\uparrow$ & codec train KL $\downarrow$ \\
    \midrule
    teacher & n/a & 0.820 & n/a \\
    LLA $4.00\times$  & $(96,160)$  & 0.590 & 0.0160 \\
    LLA $3.20\times$  & $(120,200)$ & 0.575 & 0.0153 \\
    LLA $2.67\times$  & $(144,240)$ & 0.575 & 0.0157 \\
    LLA $2.00\times$  & $(192,320)$ & \textbf{0.750} & 0.0155 \\
    LLA $1.60\times$  & $(240,400)$ & 0.755 & 0.0070 \\
    LLA $1.33\times$  & $(288,480)$ & \textbf{0.795} & 0.0035 \\
    \bottomrule
  \end{tabular}
\end{table}%
}

\newcommand{\TableBroadSuite}{%
\begin{table}[H]
  \centering\footnotesize
  \caption{\textbf{Quality against the Ouro-1.4B teacher across the broad suite.} GSM8K is 5-shot strict EM; HumanEval and MBPP use EvalPlus pass@1 (base/plus); BBH and MMLU-Pro use exact match. MATH-500 uses reliable final-answer extraction and symbolic verification. Light compression matches the teacher across tasks, while heavier compression is task-dependent.}
  \label{tab:broadsuite}
  \resizebox{\textwidth}{!}{%
  \begin{tabular}{lcccccc}
    \toprule
    Model & GSM8K & HumanEval b/p & MBPP b/p & MATH-500 & BBH & MMLU-Pro \\
    \midrule
    Ouro teacher $(1\times)$ & 0.794 & 0.707 / 0.671 & 0.720 / 0.595 & 0.74 & 0.708 & 0.481 \\
    LLA $1.33\times$ & 0.795 & 0.756 / 0.713 & 0.725 / 0.606 & 0.70 & 0.705 & 0.467 \\
    LLA $1.60\times$ & 0.755 & 0.726 / 0.689 & 0.733 / 0.624 & 0.67 & 0.695 & 0.475 \\
    LLA $2.00\times$ & 0.750 & 0.744 / 0.695 & 0.733 / 0.603 & 0.63 & 0.677 & 0.469 \\
    LLA $2.67\times$ & 0.575 & 0.543 / 0.488 & 0.672 / 0.566 & 0.54 & 0.578 & 0.402 \\
    LLA $3.20\times$ & 0.575 & 0.591 / 0.537 & 0.646 / 0.537 & 0.72 & 0.495 & 0.399 \\
    LLA $4.00\times$ & 0.590 & 0.610 / 0.561 & 0.616 / 0.513 & 0.52 & 0.527 & 0.416 \\
    \bottomrule
  \end{tabular}}
\end{table}%
}

\newcommand{\FigureBroadSuite}{%
\begin{figure}[t]
  \centering
  \includegraphics[width=0.8\linewidth]{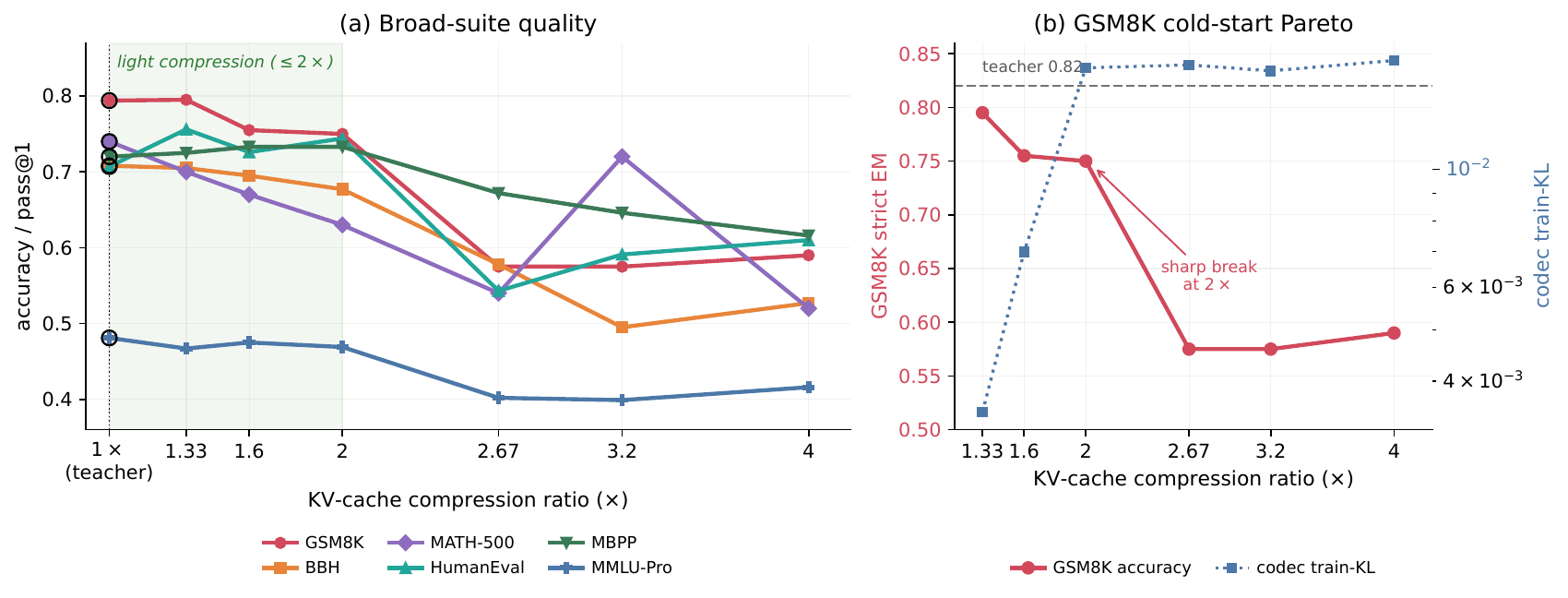}
  \caption{\textbf{Broad-suite quality against the Ouro-1.4B teacher as KV-cache compression increases.} Effects are task-dependent: code (HumanEval/MBPP) and MMLU-Pro stay at or above the teacher through $4\times$, while the reasoning
  tasks degrade - only gently within the light-compression region ($\le 2\times$, shaded) for GSM8K and BBH, but throughout for the more sensitive MATH-500, and sharply beyond $2\times$ for GSM8K and BBH. GSM8K is 5-shot strict EM HumanEval/MBPP are EvalPlus pass@1 (base); MATH-500 uses symbolic final-answer verification; BBH and MMLU-Pro use exact match. Full numbers are in Table~\ref{tab:broadsuite}.}
  \label{fig:broadsuite}
\end{figure}%
}

\newcommand{\TableIsoCache}{%
\begin{table}[H]
  \centering\footnotesize
  \caption{\textbf{Iso-cache comparison across compression axes.}}
  \label{tab:isocache}
  \vspace{0.5em}
  \begin{tabular}{llcccc}
    \toprule
    Method & $\rho$ & train-KL $\downarrow$ & GSM8K $\uparrow$ & MC-avg $\uparrow$ & MMLU $\uparrow$ \\
    \midrule
    teacher & $1\times$ & n/a & 0.794 & 0.646 & 0.682 \\
    \midrule
    final-loop reuse & $4\times$ & n/a & 0.000 & 0.443 & 0.537 \\
    \rowcolor{llahl} LLA (loop) & $4\times$ & \textbf{0.059} & \textbf{0.800} & 0.640 & \textbf{0.672} \\
    mla\_head (head) & $4\times$ & 0.122 & 0.522 & 0.585 & 0.657 \\
    cla (layer) & $4\times$ & 0.267 & 0.660 & 0.597 & 0.666 \\
    kv\_quant (precision) & $4\times$ & n/a & 0.702 & \textbf{0.645} & 0.672 \\
    \midrule
    \rowcolor{llahl} LLA (loop) & $5.33\times$ & 0.062 & 0.752 & \textbf{0.640} & \textbf{0.674} \\
    mla\_head & $5.33\times$ & 0.123 & 0.180 & 0.576 & 0.640 \\
    LLA-2D (loop$\times$head) & $5.33\times$ & 0.078 & \textbf{0.760} & 0.628 & \textbf{0.674} \\
    kv\_quant (precision) & $5.33\times$ & n/a & 0.122 & 0.621 & 0.607 \\
    \midrule
    \rowcolor{llahl} LLA (loop) & $21.33\times$ & 0.128 & \textbf{0.740} & 0.622 & 0.668 \\
    mla\_head & $21.33\times$ & 0.128 & 0.680 & 0.613 & 0.633 \\
    LLA-2D (loop$\times$head) & $21.33\times$ & \textbf{0.097} & 0.723 & \textbf{0.624} & \textbf{0.670} \\
    \bottomrule
  \end{tabular}
\end{table}%
}

\newcommand{\TableScale}{%
\begin{table}[H]
  \centering\footnotesize
  \caption{\textbf{Scale generalization on Ouro-2.6B-Thinking.} GSM8K is evaluated with the cached reconstruction decoder, while train-KL is decoder-independent. LLA remains near-lossless across $4$ to $10.67\times$, while the head and layer baselines lose substantially more accuracy.}
  \label{tab:scale}
  \begin{tabular}{lcccc}
    \toprule
    $\rho$ & LLA (loop) & mla\_head & LLA-2D & kv\_quant \\
    \midrule
    \multicolumn{5}{c}{GSM8K EM $\uparrow$; teacher $\approx 0.83$} \\
    $4\times$ & \textbf{0.834} & 0.585 & 0.66 & 0.814 \\
    $5.33\times$ & \textbf{0.817} & 0.772 & 0.74 & 0.351 \\
    $10.67\times$ & \textbf{0.821} & 0.775 & 0.82 & n/a \\
    \midrule
    \multicolumn{5}{c}{train-KL $\downarrow$} \\
    $4\times$ & \textbf{0.057} & 0.150 & 0.192 & n/a \\
    $5.33\times$ & \textbf{0.056} & 0.155 & 0.165 & n/a \\
    $10.67\times$ & \textbf{0.070} & 0.121 & 0.110 & n/a \\
    \bottomrule
  \end{tabular}
\end{table}%
}

\newcommand{\TableOnPolicy}{%
\begin{table}[H]
  \centering\footnotesize
  \caption{\textbf{On-policy distillation across the rank Pareto.} MATH-500 is scored by reliable final-answer extraction plus symbolic verification (teacher $0.74$). On-policy training improves long-generation stability at every ratio and gives the largest gains at $3.2$ to $5.33\times$. ``no-ans'' denotes no extractable final answer.}
  \label{tab:onpolicy}
  \begin{tabular}{lccccc}
    \toprule
    $\rho$ & off-policy & on-policy & $\Delta$ & off no-ans & on no-ans \\
    \midrule
    teacher $(1\times)$ & 0.740 & n/a & n/a & 0.020 & n/a \\
    $3.2\times$ & 0.460 & \textbf{0.700} & $+0.24$ & 0.21 & 0.06 \\
    $4\times$ & 0.430 & \textbf{0.660} & $+0.23$ & 0.15 & 0.03 \\
    $5.33\times$ & 0.530 & \textbf{0.690} & $+0.16$ & 0.12 & 0.03 \\
    $10.67\times$ & 0.650 & \textbf{0.690} & $+0.04$ & 0.09 & 0.07 \\
    $21.33\times$ & \textbf{0.660} & 0.620 & $-0.04$ & 0.11 & 0.05 \\
    \bottomrule
  \end{tabular}
\end{table}%
}

\newcommand{\TableAblations}{%
\begin{table}[H]
  \centering\footnotesize
  \caption{\textbf{Design-choice ablations.} Held-out teacher-forced KL for a $4\times$ codec after a matched 300-step conversion. The reference is SVD initialization, split ranks $r_k/r_v=96/160$ and attention-output matching.}
  \label{tab:ablations}
  \begin{tabular}{llc}
    \toprule
    Ablation & Variant & held-out KL $\downarrow$ \\
    \midrule
    init prior & SVD (ours) & \textbf{0.105} \\
    init prior & $t=1$ SVD & 0.128 \\
    init prior & random & 0.771 \\
    \midrule
    rank split & $r_k<r_v$ $(96/160)$ & \textbf{0.105} \\
    rank split & shared rank $(128/128)$ & 0.129 \\
    \midrule
    loss & KL $+\lambda_{\rm attn}=0.5$ & \textbf{0.105} \\
    loss & KL only & 0.110 \\
    \bottomrule
  \end{tabular}
\end{table}%
}

\newcommand{\TableHuginn}{%
\begin{table}[H]
  \centering\footnotesize
  \caption{\textbf{LLA on Huginn-3.5B with $R=32$ recurrence steps.} Decoder-independent fidelity is measured against the uncompressed model on held-out text. A training-free SVD codec is near-lossless to $32\times$, and short KL distillation recovers the $64$ to $128\times$ range. MC-avg is a broad downstream loglikelihood suite.}
  \label{tab:huginn}
  \begin{tabular}{lcccccc}
    \toprule
    $\rho$ & $r_k/r_v$ & SVD KL $\downarrow$ & SVD top-1 & distill KL $\downarrow$ & best top-1 & MC-avg \\
    \midrule
    teacher & n/a & n/a & n/a & n/a & n/a & 0.523 \\
    $16\times$ & $128/256$ & 0.005 & 0.970 & n/a & 0.970 & 0.523 \\
    $32\times$ & $64/128$ & 0.010 & 0.954 & n/a & 0.954 & 0.525 \\
    $64\times$ & $32/64$ & 0.197 & 0.799 & 0.145 & 0.834 & 0.486 \\
    $128\times$ & $16/32$ & 0.763 & 0.620 & 0.177 & 0.812 & 0.473 \\
    \bottomrule
  \end{tabular}
\end{table}%
}

\newcommand{\TableMemory}{%
\begin{table}[H]
  \centering\footnotesize
  \caption{\textbf{Memory, capacity and latency.} LLA reduces cached scalars exactly by $\rho$. At fixed VRAM and context length 4096, measured maximum batch follows the expected capacity gain. The full-memory fast path caches reconstructed K/V and is fast, while the latent-store capacity path realizes the memory saving and is reconstruction-compute-bound in the current implementation. Combining $21.3\times$ with token eviction reduces the 1M-context 1.4B cache further to $36$~MiB. ``n/a'' marks an operating point not characterized for that path rather than a missing measurement. The fast path is profiled at its intended $4\times$ point (it caches full K/V and does not realize the memory saving at higher $\rho$), the capacity path at $21.3\times$, and the 2.6B capacity test was run to $4\times$.}
  \label{tab:memory}
  \resizebox{0.85\textwidth}{!}{%
  \begin{tabular}{lccc}
    \toprule
    Quantity & teacher & LLA $4\times$ & LLA $21.3\times$ \\
    \midrule
    KV / token, 1.4B / 2.6B & 768 / 1536 KiB & 192 / 384 KiB & 36 / 72 KiB \\
    KV at 1M context, 1.4B & 768 GiB & 192 GiB & 36 GiB \\
    max batch at 4k, 1.4B & 32 & 128 & 768 \\
    max batch at 4k, 2.6B & 16 & 64 & n/a \\
    decode latency, fast path & 63.6 ms/tok & 74.1 ms/tok & n/a \\
    peak decode throughput, capacity path & 748 tok/s & n/a & 108 tok/s \\
    \bottomrule
  \end{tabular}}
\end{table}%
}

\newcommand{\TableEvict}{%
\begin{table}[H]
  \centering\footnotesize
  \caption{\textbf{LLA composes with token eviction at no extra quality cost.} GSM8K
  (5-shot, strict EM) under StreamingLLM/H2O-style eviction to a $260$-token
  budget (sink $4$ + window $256$). The $\sim$1k-token 5-shot prompts are $\sim$74\%
  evicted, so the sequence-axis cost is real (both rows fall from $\approx0.79$ without
  eviction; teacher/LLA-$4\times$ no-eviction reference in Table~\ref{tab:isocache}).
  Eviction on a near-lossless 8-bit codec isolates the sequence axis, and LLA-$4\times$ adds
  the recurrence axis on top. The two remain close, confirming that the axes are orthogonal.
  LLA reconstructs the \emph{kept} tokens accurately and adds little beyond eviction's own cost.
  Cache size composes exactly (Table~\ref{tab:memory}). LLA-$21.3\times$ $+$ eviction caps
  the 1M-context 1.4B cache at $36$~MiB.}
  \label{tab:evict}
  \begin{tabular}{lc}
    \toprule
    Method ($260$-token eviction budget) & GSM8K strict $\uparrow$ \\
    \midrule
    eviction alone (kv\_quant-8, sequence axis) & 0.510 \\
    LLA $4\times$ $+$ eviction (loop $\times$ sequence) & 0.520 \\
    \bottomrule
  \end{tabular}
\end{table}%
}

\newcommand{\FigurePasskey}{%
\begin{figure}[ht]
  \centering
  \includegraphics[width=\linewidth]{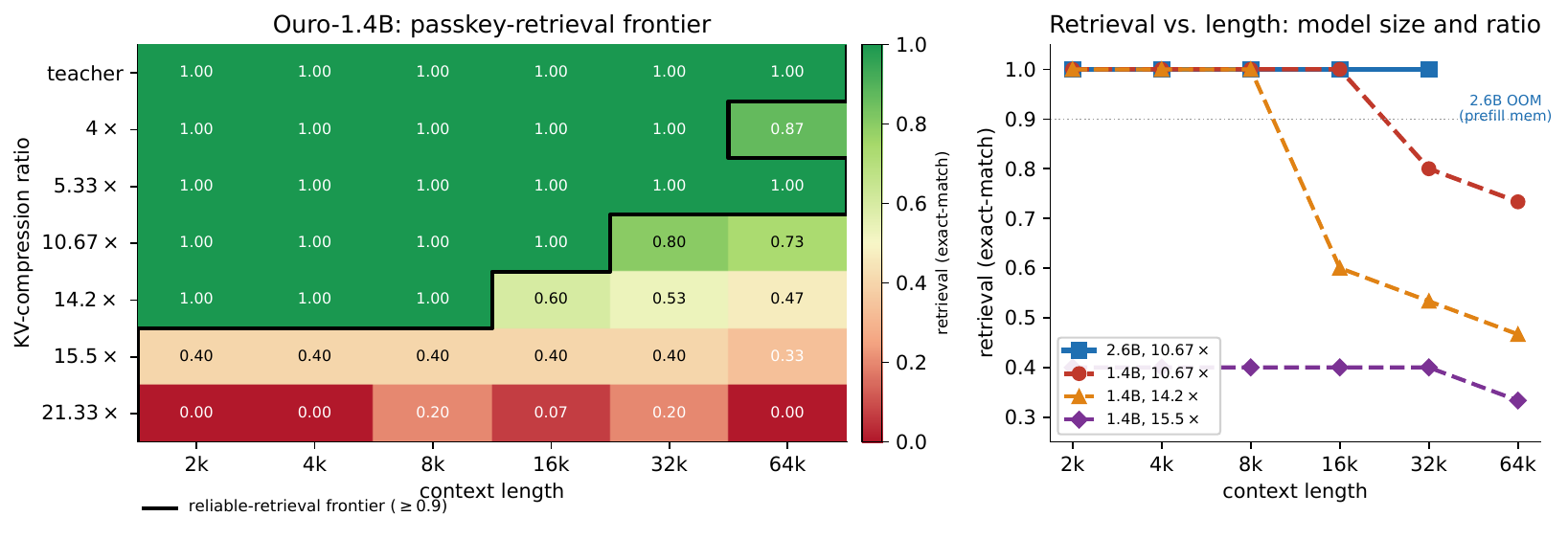}
  \caption{\textbf{Passkey retrieval under KV-cache compression.} \emph{Left:} exact
  retrieval for Ouro-1.4B over the (compression ratio $\times$ context length) grid
  ($15$ trials/cell; every cell is a measured run, full numbers in
  Table~\ref{tab:passkey}). The black staircase is the empirical
  reliable-retrieval frontier (the $\ge 0.9$ iso-retrieval contour through measured cells).
  \emph{Right:} retrieval vs.\ length for the 1.4B model at $10.67/14.2/15.5\times$ and the
  2.6B model at $10.67\times$, exposing both degradation regimes and the model-size effect in
  one view.}
  \label{fig:passkey}
\end{figure}%
}

\newcommand{\TablePasskey}{%
\begin{table}[H]
  \centering\footnotesize
  \caption{\textbf{Long-context passkey retrieval under LLA compression (Ouro-1.4B).}
  Exact-match retrieval of a 5-digit key hidden at a random depth in filler context.
  The uncompressed teacher retrieves perfectly to $64$k. LLA is near-lossless for retrieval
  through $5.33\times$ across the whole range, with only the $4\times$, $64$k cell dipping
  to $0.87$. The $10.67\times$ point is the knee (perfect to $16$k, degrading at
  $32$ to $64$k), while $21.33\times$ loses the exact token identity a key requires.
  LLA rows use the capacity/latent cached decoder, whose two-pass prefill reaches $64$k
  via streamed layer-wise reconstruction (Sec.~\ref{sec:systems}).}
  \label{tab:passkey}
  \begin{tabular}{rccccc}
    \toprule
    context & teacher & LLA $4\times$ & LLA $5.33\times$ & LLA $10.67\times$ & LLA $21.33\times$ \\
    \midrule
    2{,}048  & 1.00 & 1.00 & 1.00 & 1.00 & 0.00 \\
    4{,}096  & 1.00 & 1.00 & 1.00 & 1.00 & 0.00 \\
    8{,}192  & 1.00 & 1.00 & 1.00 & 1.00 & 0.20 \\
    16{,}384 & 1.00 & 1.00 & 1.00 & 1.00 & 0.07 \\
    32{,}768 & 1.00 & 1.00 & 1.00 & 0.80 & 0.20 \\
    65{,}536 & 1.00 & 0.87 & 1.00 & 0.73 & 0.00 \\
    \bottomrule
  \end{tabular}
\end{table}%
}

\newcommand{\TableScaleMC}{%
\begin{table}[H]
  \centering\scriptsize
  \caption{\textbf{Ouro-2.6B-Thinking: full commonsense and knowledge suite.} All tasks are 0-shot loglikelihood accuracy and decoder-independent. Per-head LLA is near-lossless at every budget and remains several points above the head and layer baselines and the LLA-2D variant.}
  \label{tab:scale_mc}
  \resizebox{\textwidth}{!}{%
  \begin{tabular}{llccccccccc}
    \toprule
    Method & $\rho$ & ARC-e & ARC-c & HSwag & WGr & C-QA & MMLU & PIQA & OBQA & Avg \\
    \midrule
    teacher & $1\times$ & 0.833 & 0.541 & 0.568 & 0.684 & 0.752 & 0.750 & 0.785 & 0.310 & 0.653 \\
    \rowcolor{llahl} LLA & $4\times$ & 0.818 & 0.521 & 0.557 & 0.676 & 0.754 & 0.749 & 0.782 & 0.302 & 0.645 \\
    \rowcolor{llahl} LLA & $5.33\times$ & 0.818 & 0.526 & 0.557 & 0.691 & 0.746 & 0.748 & 0.783 & 0.300 & 0.646 \\
    \rowcolor{llahl} LLA & $10.67\times$ & 0.813 & 0.526 & 0.552 & 0.676 & 0.741 & 0.745 & 0.780 & 0.288 & 0.640 \\
    mla\_head & $4\times$ & 0.733 & 0.460 & 0.494 & 0.681 & 0.723 & 0.740 & 0.729 & 0.234 & 0.599 \\
    mla\_head & $5.33\times$ & 0.751 & 0.474 & 0.504 & 0.660 & 0.731 & 0.747 & 0.739 & 0.244 & 0.606 \\
    mla\_head & $10.67\times$ & 0.730 & 0.461 & 0.498 & 0.668 & 0.726 & 0.747 & 0.725 & 0.256 & 0.601 \\
    LLA-2D & $4\times$ & 0.738 & 0.463 & 0.513 & 0.661 & 0.753 & 0.741 & 0.742 & 0.242 & 0.607 \\
    LLA-2D & $5.33\times$ & 0.753 & 0.477 & 0.520 & 0.664 & 0.740 & 0.740 & 0.760 & 0.236 & 0.611 \\
    LLA-2D & $10.67\times$ & 0.786 & 0.512 & 0.539 & 0.672 & 0.745 & 0.742 & 0.761 & 0.294 & 0.631 \\
    cla & $4\times$ & 0.756 & 0.484 & 0.518 & 0.676 & 0.647 & 0.725 & 0.750 & 0.260 & 0.602 \\
    cla & $6\times$ & 0.769 & 0.493 & 0.505 & 0.672 & 0.640 & 0.692 & 0.739 & 0.254 & 0.596 \\
    cla & $12\times$ & 0.732 & 0.468 & 0.484 & 0.669 & 0.595 & 0.684 & 0.717 & 0.250 & 0.575 \\
    kv\_quant & $4\times$ & 0.832 & 0.530 & 0.565 & 0.677 & 0.738 & 0.747 & 0.782 & 0.292 & 0.645 \\
    kv\_quant & $5.33\times$ & 0.816 & 0.519 & 0.550 & 0.663 & 0.712 & 0.719 & 0.777 & 0.292 & 0.631 \\
    \bottomrule
  \end{tabular}}
\end{table}%
}

\newcommand{\FigureParetoOne}{%
\begin{figure}[!htbp]
  \centering
  \includegraphics[width=0.93\textwidth]{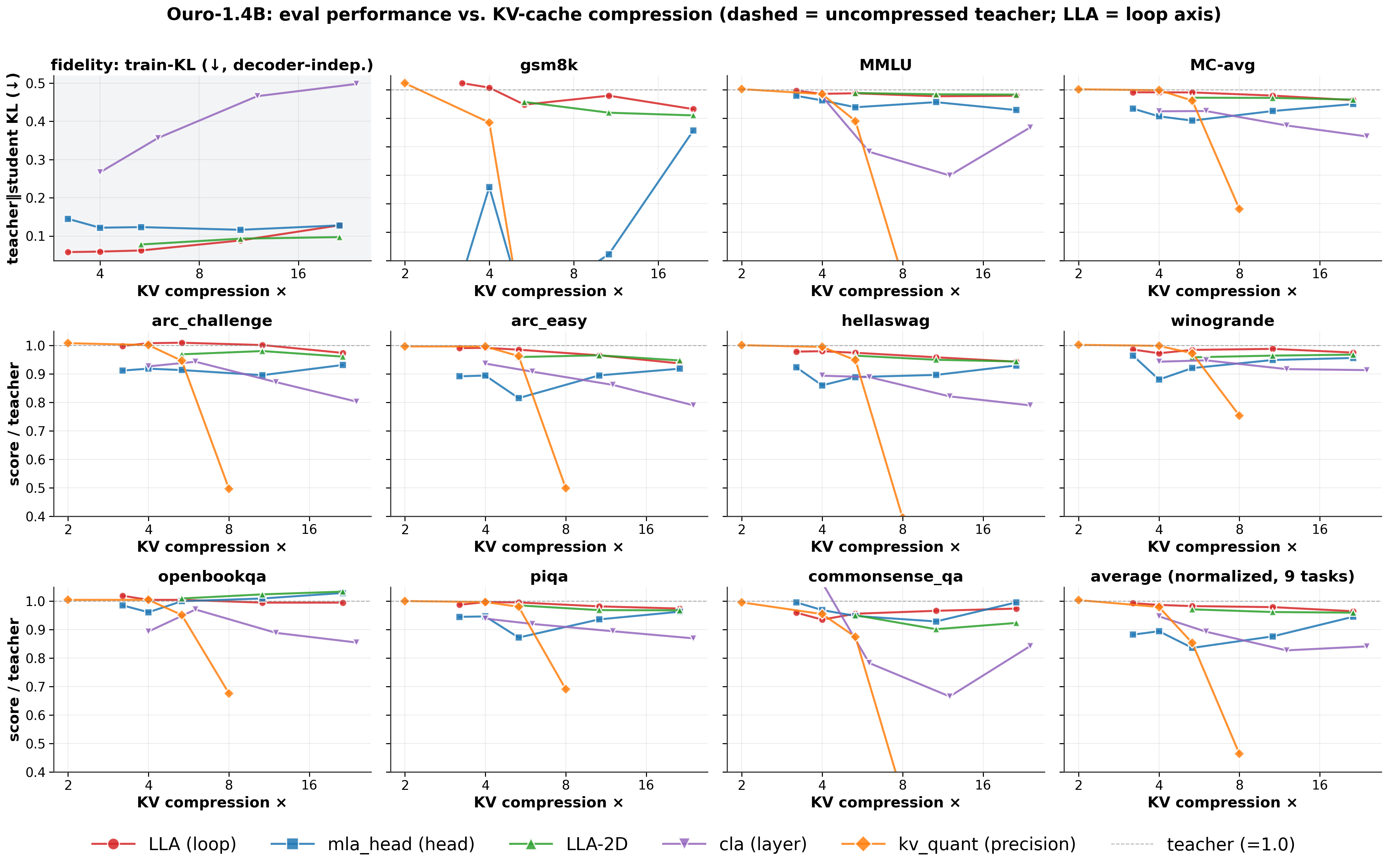}
  \caption{\textbf{Ouro-1.4B performance versus KV-cache compression.} Each panel plots an evaluation metric against compression for per-head LLA, the LLA-2D (loop$\times$head) variant, and the head, layer and precision baselines. The dashed line is the uncompressed teacher. LLA tracks the teacher most closely across the suite, while precision quantization collapses on GSM8K past moderate compression.}
  \label{fig:pareto1p4b}
\end{figure}%
}

\newcommand{\FigureParetoTwoSix}{%
\begin{figure}[!htbp]
  \centering
  \includegraphics[width=0.93\textwidth]{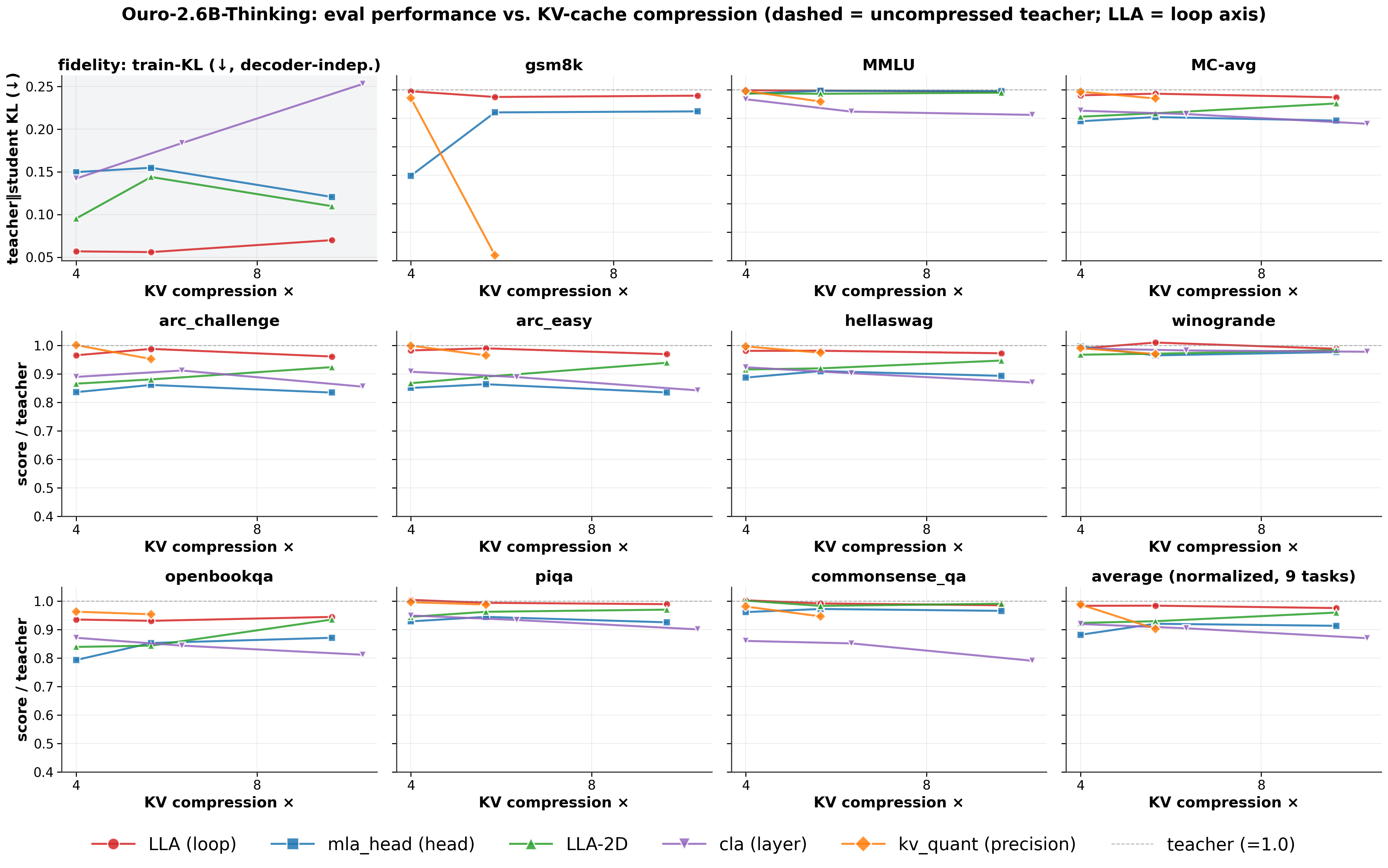}
  \caption{\textbf{Ouro-2.6B-Thinking performance versus KV-cache compression.} The layout matches Fig.~\ref{fig:pareto1p4b}. The loop-axis codec remains closest to the teacher on train-KL, GSM8K and the commonsense/knowledge suite. The precision baseline cannot reach the largest ratios without collapse.}
  \label{fig:pareto26b}
\end{figure}%
}

\newcommand{\FigureLoopRank}{%
\begin{figure}[ht]
  \centering
  \includegraphics[width=\textwidth]{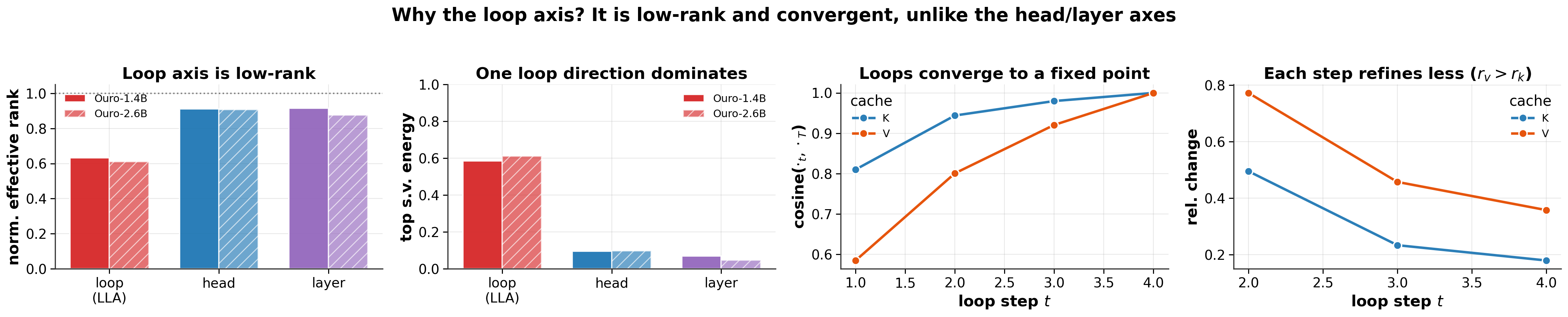}
  \caption{\textbf{Cross-loop K/V is low-rank, and its per-loop trajectory stabilizes.} Left two panels: the loop axis has much lower normalized effective rank than the head or layer axes, and a single direction carries a large fraction of the energy. Right two panels: K and V move toward the final loop while the step-to-step change shrinks, and V converges more slowly than K, motivating $r_v>r_k$.}
  \label{fig:whyloops}
\end{figure}%
}

\newcommand{\FigureTrajectory}{%
\begin{figure}[tbp]
  \centering
  \includegraphics[width=\textwidth]{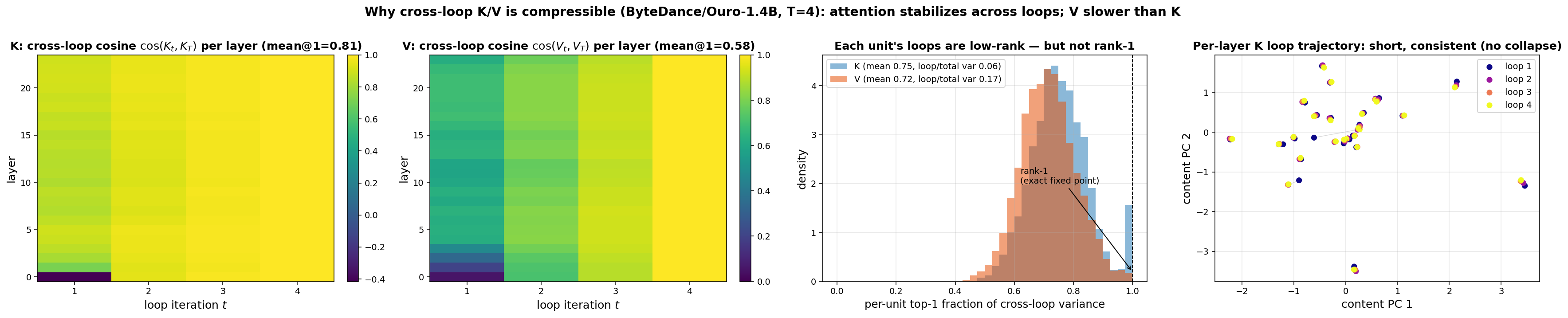}
  \caption{\textbf{Loop trajectories are low-rank, not collapsed.} In frozen Ouro-1.4B, cross-loop K/V cosines increase toward the final loop, and one direction captures most but not all cross-loop variance. The PCA view shows short, consistent paths rather than a single shared endpoint.}
  \label{fig:whytraj}
\end{figure}%
}

\newcommand{\FigureHuginn}{%
\begin{figure}[H]
  \centering
  \includegraphics[width=0.72\textwidth]{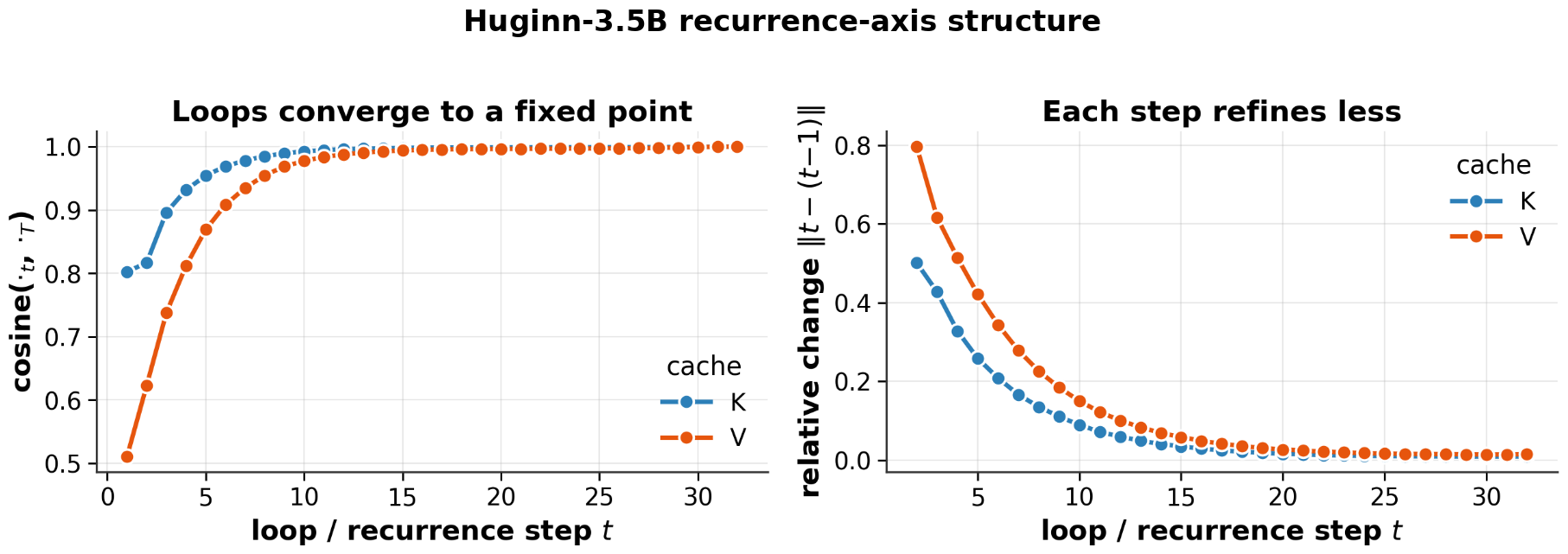}
  \caption{\textbf{The recurrence-axis mechanism transfers to Huginn-3.5B.} These are the same two convergence diagnostics as the right two panels of Fig.~\ref{fig:whyloops}, now applied to a different family. Huginn has up to 32 recurrence steps and a different looped-LM architecture, yet the pattern is unchanged. Successive steps converge toward a fixed point, each step refines less, and V converges more slowly than K. Reproducing the identical diagnostic at 32 steps, far beyond Ouro's 4, is stronger evidence for this mechanism.}
  \label{fig:whyhuginn}
\end{figure}%
}

\newcommand{\FigureAblationInit}{%
\begin{figure}[tbp]
  \centering
  \includegraphics[width=0.6\textwidth]{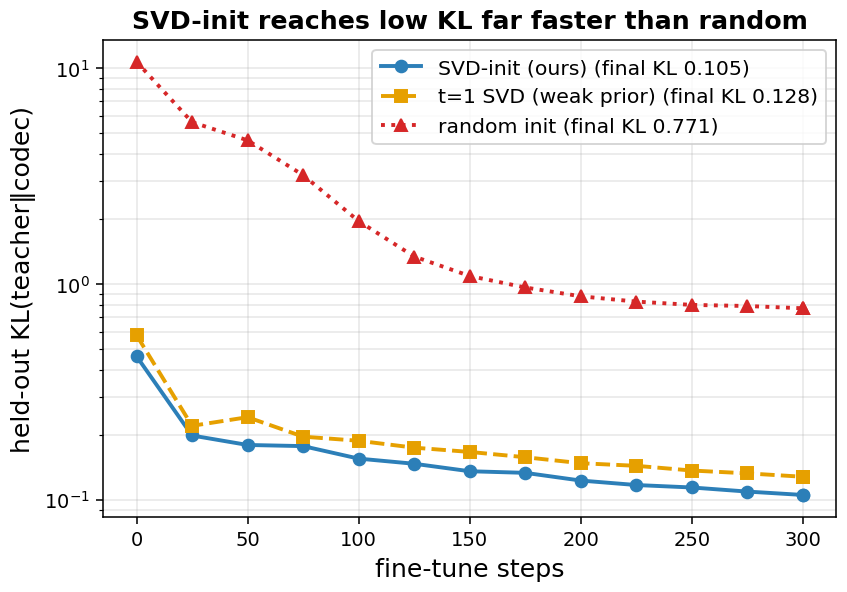}
  \caption{\textbf{SVD initialization dominates random initialization.} Held-out KL for the $4\times$ codec falls quickly from the SVD teacher prior. A random codec does not match SVD initialization within the same conversion budget.}
  \label{fig:ablinit}
\end{figure}%
}

\newcommand{\FigureOnPolicy}{%
\begin{figure}[!htbp]
  \centering
  \includegraphics[width=0.88\textwidth]{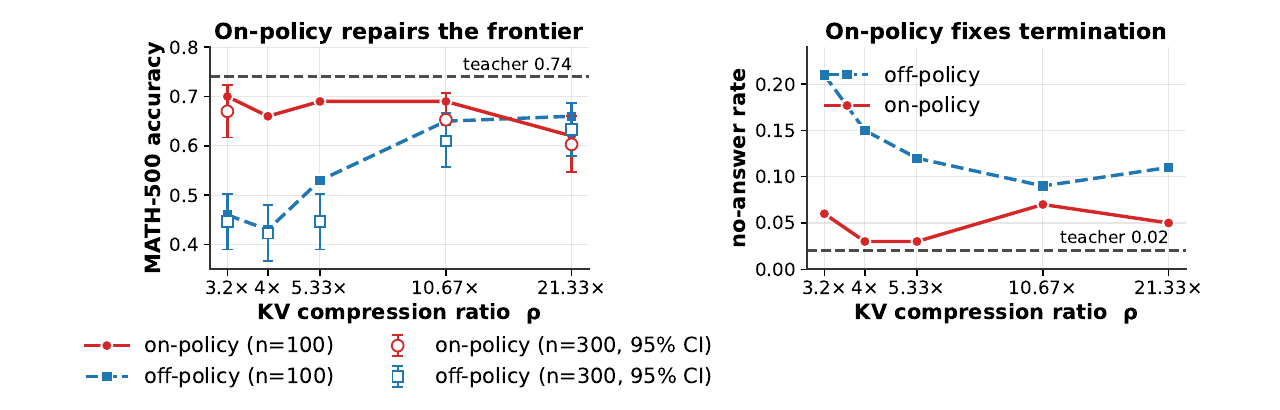}
  \caption{\textbf{On-policy distillation stabilizes the long-generation Pareto.} On reliably scored MATH-500, self-rollout distillation raises the low-compression points and reduces no-answer generations throughout the Pareto. Hollow markers show independent re-evaluations with bootstrap 95\% confidence intervals.}
  \label{fig:onpolicy}
\end{figure}%
}

\newcommand{\FigureExposure}{%
\begin{figure}[!htbp]
  \centering
  \includegraphics[width=0.88\textwidth]{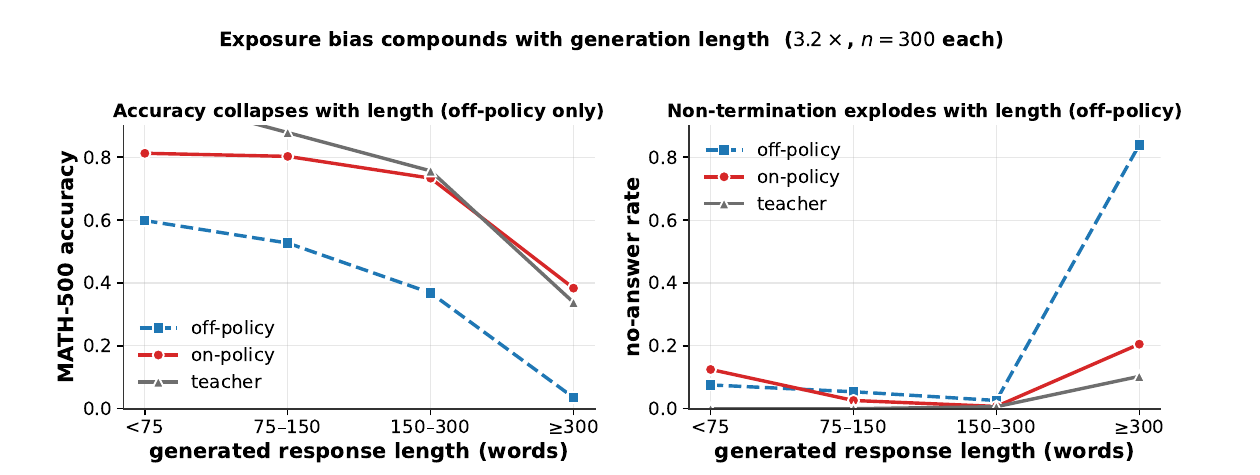}
  \caption{\textbf{Off-policy failure compounds with generation length.} At $3.2\times$, the uncompressed teacher (gray) sets the reference. Accuracy declines gently with length from problem difficulty, and non-termination stays low. On-policy distillation (red) tracks the teacher closely at every length, whereas the off-policy codec (blue) loses accuracy as responses grow longer and its no-answer rate spikes in the longest bin. The gap between off-policy and the teacher is the exposure-bias failure that on-policy training removes.}
  \label{fig:exposure}
\end{figure}%
}

\newcommand{\FigureCapacity}{%
\begin{figure}[!htbp]
  \centering
  \includegraphics[width=0.93\textwidth]{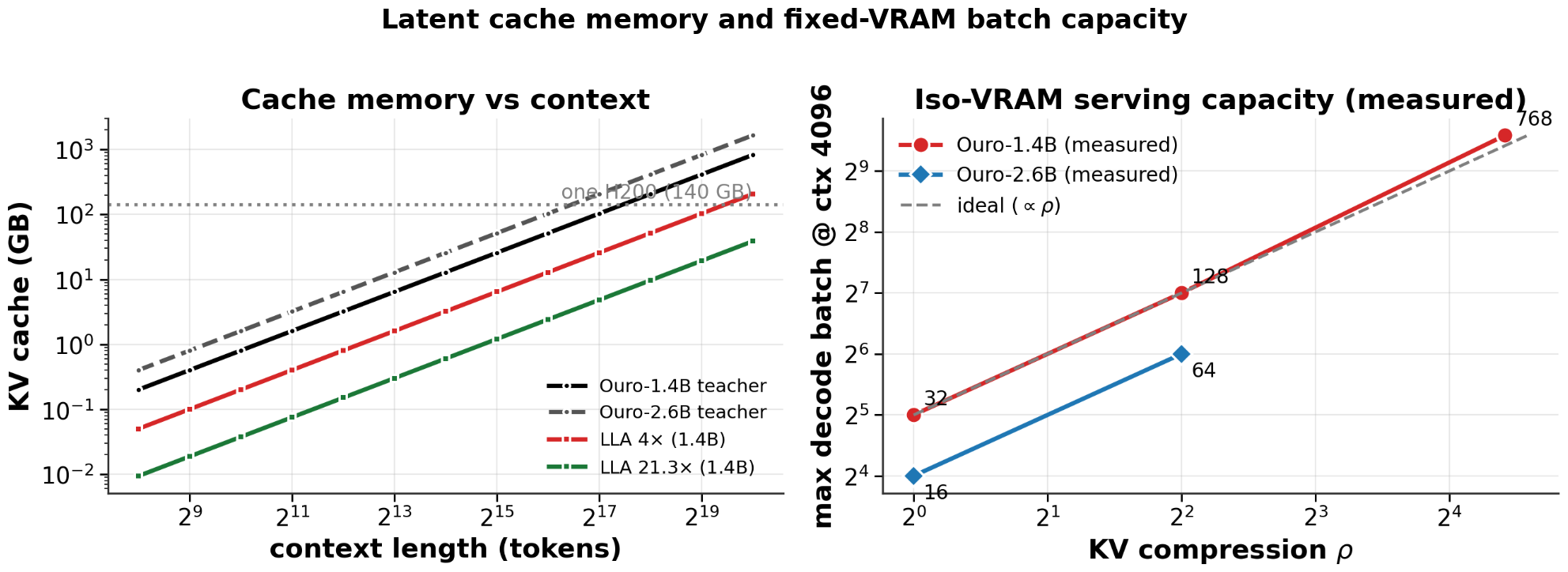}
  \caption{\textbf{The latent cache gives an exact memory reduction and measured serving capacity.} Left: cache memory scales linearly with context and is divided by the compression ratio. Right: at fixed VRAM and context length 4096, maximum batch size follows the ideal $\rho$ scaling in measured capacity tests.}
  \label{fig:capacity}
\end{figure}%
}

\newcommand{\FigureTrajectoryTwoSix}{%
\begin{figure}[H]
  \centering
  \includegraphics[width=0.75\textwidth]{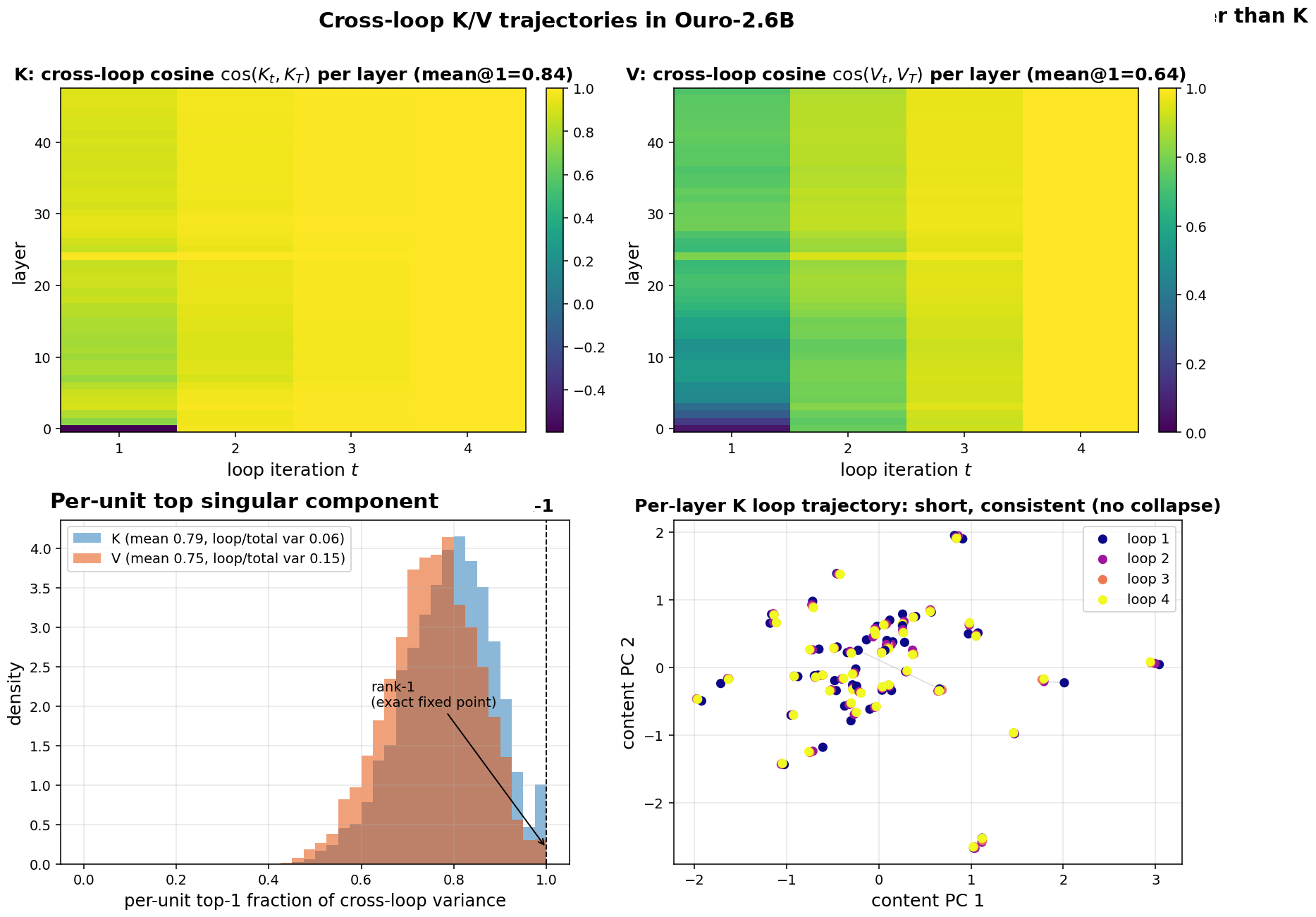}
  \caption{\textbf{Cross-loop K/V stabilization on Ouro-2.6B-Thinking.} The 2.6B model reproduces the same pattern as Ouro-1.4B. K/V stabilizes across loops, one direction dominates but does not explain all variance, and the value cache has higher cross-loop variance than the key cache.}
  \label{fig:whytraj26b}
\end{figure}%
}

\newcommand{\AppendixExpDetails}{%
\section{Experimental details}
\label{app:expdetails}

\paragraph{Models.} All experiments use frozen, publicly released looped/recurrent-depth
Transformers. \textbf{Ouro-1.4B} (the main teacher): $D=24$ weight-tied layers, $H=16$
attention heads of width $d_{\rm head}=128$, run at $T=4$ recurrence steps. \textbf{Ouro-2.6B-Thinking}:
deeper Ouro variant, also $T=4$, used for the scale study. \textbf{Huginn-3.5B}: a different
recurrent-depth family, evaluated at $R=32$ recurrence steps for the family-transfer study.
The teacher weights are never updated; only the K/V codec is trained.

\paragraph{LLA codec and conversion.} For each layer, head and axis (K, V) we collect the
teacher's pre-RoPE per-loop activations on a fixed calibration set of $255$ prompts and
initialize $W_{\rm down}$ from the top-$r$ right singular vectors of the stacked-loop matrix
(loop slices give the initial $W_{{\rm up},t}$). The SVD-initialized codec is then fine-tuned
with the teacher frozen using \texttt{train\_rmla.py}: objective = forward KL(teacher $\Vert$
swapped) $+\;\lambda_{\rm attn}\!=\!0.5$ per-loop attention-output matching (Eq.~\ref{eq:loss});
$3000$ steps ($300$ for the design ablations of Table~\ref{tab:ablations}), batch size $8$,
sequence length $512$, AdamW, learning rate $1\mathrm{e}{-3}$ with $100$-step warmup and cosine
decay to $5\%$, held-out KL evaluated every $200$ steps ($n=32$). We select the
\emph{best-KL} checkpoint (not the final step); a codec whose final-step KL exceeds its best is
reported at its best checkpoint. Ranks $(r_k,r_v)$ per compression $\rho$ are listed in
Table~\ref{tab:tier4}; we use $r_v>r_k$ throughout (V converges more slowly, Sec.~\ref{sec:structure}).
\textbf{LLA-2D} shares the same recipe but stores one latent per token and layer, reconstructing
all $H$ heads from it. RoPE is applied \emph{after} reconstruction (the rotary phase is shared
across loops for a token), so no decoupled-RoPE branch is needed for the content path.

\paragraph{Matched-axis baselines.} All learned baselines use the identical conversion recipe
above; only the codec architecture changes. \emph{mla\_head}: per-loop MLA (head-axis latent,
no cross-loop sharing). \emph{cla}: cross-layer K/V sharing over $s\in\{4,6,12,24\}$ adjacent
layers. \emph{kv\_quant}: inference-only low-bit K/V quantization at $\{2,3,4,8\}$ bits (no codec
training, hence no train-KL, and it cannot exceed $\sim$8$\times$). \emph{final-loop reuse}: a
zero-parameter control that caches only the last loop's K/V and reuses it for all loops.

\paragraph{Evaluation suite and protocols.} \textbf{GSM8K}: 5-shot at 1.4B and 2-shot at 2.6B,
strict-match EM, \texttt{max\_gen\_toks}=256. The 1.4B teacher is scored on the full $1319$-item
test set; the iso-cache and broad-suite codecs use a $100$-item subset with the faithful re-forward
decoder, while the tier4 cold-start sweep (Table~\ref{tab:tier4}) uses a $200$-item subset with the
earlier streaming reconstruction decoder and a $100$-item teacher reference. At 2.6B, scored with the
faithful decoder, the teacher and the per-head and head-axis arms use the full $1319$ items and the
\lla{}-2D arms use a $100$-item subset. \textbf{MATH-500}: 4-shot, \texttt{max\_gen\_toks}=2048,
batch 1, faithful decoder, scored by \texttt{math\_verify} symbolic equivalence; teacher on all
$500$ items, codecs on $200$. \textbf{BBH}: 3-shot, \texttt{max\_gen\_toks}=1024, batch 1, exact
match, faithful decoder; teacher $25$ items/sub-task ($675$ docs) and codecs $5$/sub-task ($135$
docs, identical items across arms). \textbf{HumanEval}/\textbf{MBPP}: EvalPlus pass@1 (base/plus)
on the full sets ($164$/$378$). \textbf{MMLU-Pro}: exact match on the full set. The
commonsense/knowledge MC suite (\textbf{ARC-e/c, HellaSwag, PIQA, WinoGrande, OpenBookQA,
CommonsenseQA, MMLU}) is 0-shot loglikelihood accuracy on the full sets and is decoder-independent.
Within each table the codec and baseline arms are scored on identical items; the teacher reference
is scored on the full set where the two differ.
Passkey (Table~\ref{tab:passkey}) hides a 5-digit key at a random depth in filler context, $15$
trials/cell ($5$ keys $\times$ $3$ depths), scored by exact retrieval; contexts $2$k to $64$k.
``train-KL'' is the held-out teacher-forced KL of the codec and is decoder-independent.

\paragraph{Decoding and caching.} Two decoders are used. The \emph{cached reconstruction decoder}
is two-pass: pass~1 captures the teacher's pre-RoPE per-loop K/V, the codec reconstructs them,
and pass~2 re-forwards forcing the reconstruction so the model's own RoPE and cache store the
post-RoPE reconstruction. This matches the training objective and is used for the iso-cache, broad-suite, 2.6B and MATH-500/BBH numbers; the tier4 cold-start sweep used an earlier streaming reconstruction decoder. The \emph{capacity /
latent-store decoder} caches only the latent and reconstructs on read, realizing the $\rho\times$
memory saving; its two-pass prefill is streamed layer-by-layer so long prompts (to $64$k) fit in
memory. Two serving regimes (Sec.~\ref{sec:systems}): a full-memory \emph{fast path} (caches
reconstructed full K/V, teacher-latency, no memory saving) and a \emph{capacity path} (stores
latents, reconstruction-compute-bound). On-policy refinement samples the codec's own rollouts
(temperature sampling, no reward/correctness filtering) and minimizes the frozen teacher's
per-token KL to those visited states with a stop-gradient through sampling.

\paragraph{Hardware and precision.} Codecs are trained and evaluated in bfloat16. Serving-capacity
and latency measurements (Table~\ref{tab:memory}) are on a single H200 at context length $4096$;
generative evaluations use B200 GPUs (183~GB) with per-process memory fraction $0.99$
for the long-context and 2.6B runs.

\paragraph{Broad-suite accuracy relative to the teacher.}
Table~\ref{tab:retention} restates the broad suite (Table~\ref{tab:broadsuite}) as compressed$/$teacher ratios, putting every task on one scale so the task-dependence of compression is directly comparable.
\begin{table}[H]
  \centering\footnotesize
  \caption{\textbf{Broad-suite accuracy relative to the Ouro-1.4B teacher} (compressed$/$teacher, computed from Table~\ref{tab:broadsuite}). A value of $1.00$ matches the teacher. The common scale makes the task-dependence explicit. Code (HumanEval, MBPP) and MMLU-Pro retain the most under compression, while GSM8K and BBH degrade fastest.}
  \label{tab:retention}
  \begin{tabular}{lcccccccc}
    \toprule
    Model & GSM8K & HE b & HE p & MBPP b & MBPP p & MATH-500 & BBH & MMLU-Pro \\
    \midrule
    Ouro teacher $(1\times)$ & 1.00 & 1.00 & 1.00 & 1.00 & 1.00 & 1.00 & 1.00 & 1.00 \\
    \midrule
    LLA $1.33\times$ & 1.00 & 1.07 & 1.06 & 1.01 & 1.02 & 0.95 & 1.00 & 0.97 \\
    LLA $1.60\times$ & 0.95 & 1.03 & 1.03 & 1.02 & 1.05 & 0.91 & 0.98 & 0.99 \\
    LLA $2.00\times$ & 0.94 & 1.05 & 1.04 & 1.02 & 1.01 & 0.85 & 0.96 & 0.98 \\
    LLA $2.67\times$ & 0.72 & 0.77 & 0.73 & 0.93 & 0.95 & 0.73 & 0.82 & 0.84 \\
    LLA $3.20\times$ & 0.72 & 0.84 & 0.80 & 0.90 & 0.90 & 0.97 & 0.70 & 0.83 \\
    LLA $4.00\times$ & 0.74 & 0.86 & 0.84 & 0.86 & 0.86 & 0.70 & 0.74 & 0.86 \\
    \bottomrule
  \end{tabular}
\end{table}%
}

\title{Looped Latent Attention: Cross-Loop KV Compression for Looped Transformers}

\author{\name James O'Neill\thanks{Corresponding author.} \email james.oneill@intercom.io \\
\addr Fin AI Research \AND
\name Fergal Reid \\
\addr Fin AI Research}

\begin{document}
\maketitle

\begin{abstract}
Looped, weight-tied Transformers reduce parameters by reusing a single block, but decoding still stores a separate K/V cache for every recurrence step. We show that this loop-indexed cache is highly structured. For a fixed token, layer and head, K/V vectors trace a short low-rank trajectory across loops, while the head and layer axes remain much flatter. We introduce Looped Latent Attention (\lla{}), a post-training cache codec that stores compact K and V latents and reconstructs loop-specific K/V vectors only when attention reads them. The default per-head codec compresses recurrence, while \lla{}-2D also folds heads into one latent for the extreme-compression regime. The codec is initialized from the SVD of teacher activations and refined with logit and attention-output distillation. At matched cache budget, per-head \lla{} outperforms head-axis MLA, cross-layer sharing, KV quantization and final-loop reuse, showing that the recurrent cache is low-rank but not safely collapsible to a single state. The same axis advantage holds on Ouro-2.6B-Thinking and transfers to Huginn-3.5B, where an SVD codec remains near-lossless to $32\times$ compression in decoder-independent evaluation. The cache reduction is exact. On one H200, the latent-store path increases measured Ouro-1.4B batch capacity at 4k context from 32 to 768 sequences at $21.3\times$ compression. Lastly, for long reasoning rollouts such as in MATH-500, on-policy refinement on student-generated prefixes raises accuracy at $4\times$ compression from 0.43 to 0.66 and reduces no-answer generations when compared to token-level off-policy distillation.
\end{abstract}

\section{Introduction}
\label{sec:intro}
At inference, the memory that constrains a Transformer is its KV cache, not its weights. The weights are loaded once and shared across the whole batch. The cache is private to each request and grows with the context and with every generated token. Past a certain context length, this per-request state, not the parameter count, decides how many requests fit on a device and whether batched serving stays economical.

Looped, weight-tied Transformers attack the parameter side of this cost by applying one block repeatedly. Universal Transformers \citep{dehghani2019universal} recast depth as recurrence, and recent looped language models such as Ouro/LoopLM \citep{ouro2024} scale the idea to billions of parameters. However, the cache is untied. Each recurrence step recomputes attention and writes its own keys and values, so the cache is indexed by recurrence step in addition to layer, head, and token. A model with $T=4$ loops over $D=24$ layers therefore carries the KV cache of a 96-layer decoder while sharing the parameters of a 24-layer one.

This cache carries far fewer degrees of freedom than its nominal size. For a fixed token, layer, and head, the per-loop K/V vectors trace a short, low-rank trajectory, and the redundancy is specific to the loop axis: at the same budget, the head and layer spectra are much flatter. Looped Latent Attention (\lla{}) exploits this by replacing the $T$ per-loop K/V vectors with single latent and loop-specific reconstruction maps.

An important question is \textit{which axis to compress and what the KV cache must preserve when it does}. A looped decoder invites a degenerate shortcut, if recurrence merely converges to a fixed point, one could store the final loop and reuse it everywhere. It also inherits the usual Transformer options of compressing heads, sharing layers, and quantizing values. Our experiments separate these by only changing what the cache stores while the teacher is kept frozen and the attention pattern and recurrence intact. Under this controlled swap, final-loop reuse buys the same $4\times$ reduction for free yet collapses GSM8K generation to zero, while a small low-rank trajectory code holds performance at matched-budget. In the following, we list our three primary contributions.

\paragraph{Contributions.}
\begin{itemize}
  \item \lla{}, a cross-loop latent K/V codec for weight-tied looped Transformers. It stores $(r_k+r_v)H$ scalars per token per layer instead of $2THd_{\rm head}$, giving compression $\rho=2Td_{\rm head}/(r_k+r_v)$. \lla{}-2D extends the same family by folding the heads into one latent for the extreme-compression regime.
  \item We show that recurrence is the most compressible cache axis in Ouro-1.4B, Ouro-2.6B-Thinking and Huginn-3.5B. K and V follow different trajectories, motivating $r_v>r_k$, and matched-budget comparisons against head-axis MLA, cross-layer sharing, KV quantization and final-loop reuse favor the loop axis.
  \item We tie the results to deployment: an explicit on-policy refinement objective uses student-generated prefixes to stabilize long rollouts, and the exact scalar reduction translates into measured serving gains, raising Ouro-1.4B maximum batch from 32 to 768 sequences at $21.3\times$ compression.
\end{itemize}
\section{Related Work}
\label{sec:related}
\paragraph{Looped and Universal Transformers.}
Universal Transformers \citep{dehghani2019universal} apply a weight-tied block for several steps, casting depth as recurrence. Recent looped language models carry this to billion-parameter LMs and trade computation for parameter efficiency on reasoning workloads \citep{ouro2024}. Mechanistic studies of these models report that per-loop computation settles onto low-dimensional trajectories \citep{blayney2026mechanisticanalysisloopedreasoning}, and \lla{} builds a cache codec on exactly that structure. Recurrence does not address serving cost, where attention still stores a separate K/V state for every loop.

\paragraph{Latent and compressed KV caches.}
Multi-head Latent Attention (MLA) introduced in DeepSeek-V2 compresses the standard Transformer KV cache by storing a low-rank latent across the head/channel axis and reconstructing per-head keys and values \citep{deepseekv2}. Cross-layer sharing targets the layer axis \citep{cla2024,yoco2024,minicache2024}, quantization targets precision \citep{kvquant2024}, and token eviction targets the sequence axis \citep{h2o2023,streamingllm2024}. All of these operate on single-pass Transformers. \lla{} instead compresses recurrence, an axis that only weight-tied looped models have. Because the same block and projections are reused at every step, the loop axis is redundant for a reason the head and layer axes lack. The head axis, targeted by multi-query and grouped-query attention \citep{mqa2019,gqa2023}, composes on top of \lla{} (\lla{}-2D), and sequence-axis eviction multiplies with it (Sec.~\ref{sec:systems}).

\paragraph{Memory-efficient looped Transformers.}
MELT \citep{vendrell2026melt} is the closest concurrent work. It distills a single gated K/V state per layer, while \lla{} keeps the loop trajectory and compresses its rank. However, as we will see, our zero-parameter final-loop-reuse control, which collapses the loop to one state, fails at the same cache budget. The two methods also span different regimes. A collapsed state fixes the memory reduction at roughly the loop count, whereas a tunable latent can sit below or above that point by choice of $r_k$ and $r_v$.

\begin{figure}[!htbp]
  \centering
  \includegraphics[width=0.9\textwidth]{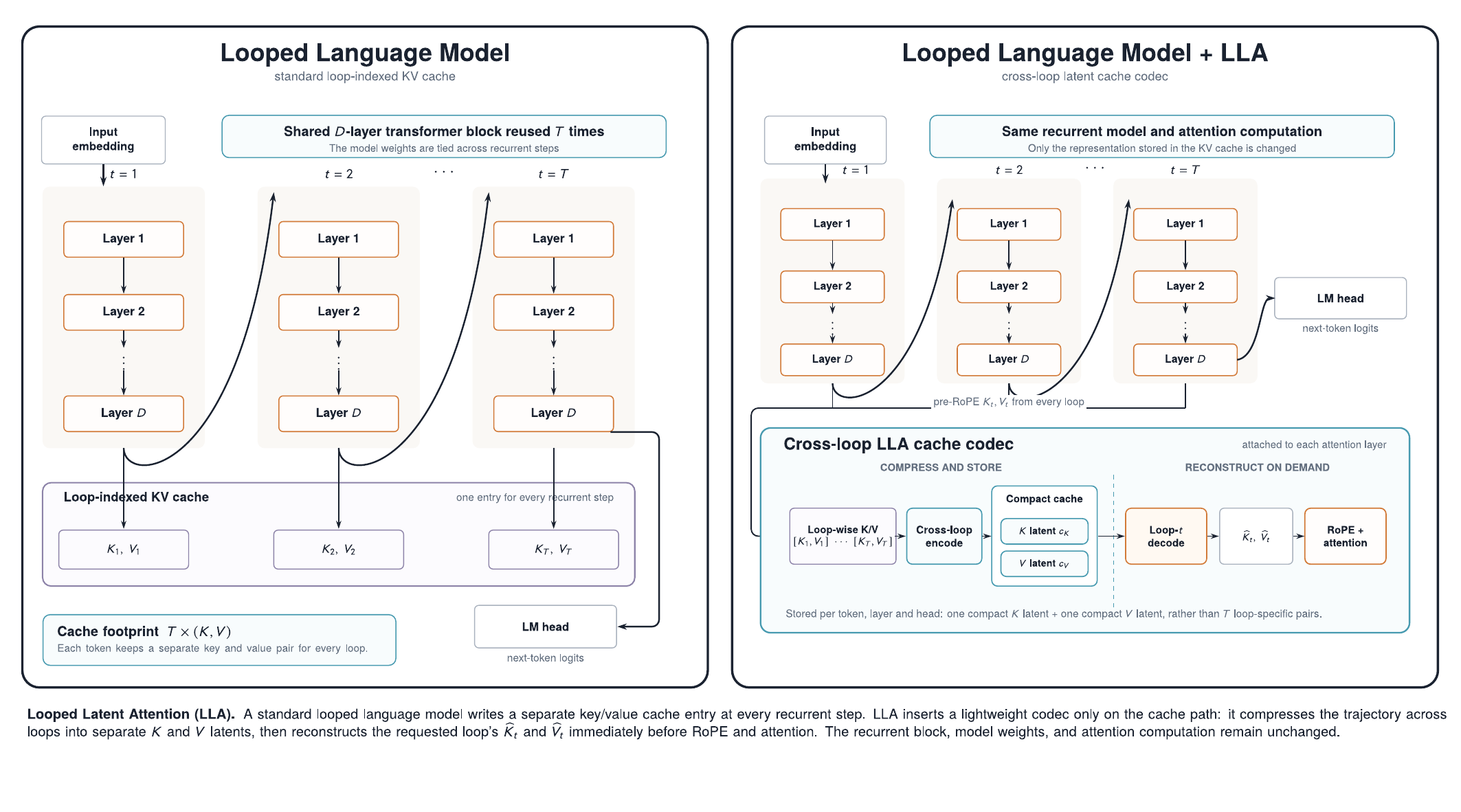}
  \vspace{-1.5em}
  \caption{\textbf{Cross-loop Looped Latent Attention.}
\textbf{Left:} A weight-tied looped language model reuses the same $D$-layer decoder stack for $T$ recurrent steps. In the standard implementation, each step contributes a separate key--value pair $(K_t,V_t)$ to a loop-indexed KV cache, so the stored cache grows with the number of recurrent steps.
\textbf{Right:} Looped Latent Attention replaces these loop-specific cache entries with a cross-loop codec. For each layer, token, and attention head, the keys and values produced across the $T$ recurrent steps are collected along the loop axis and compressed into separate compact K and V latents. At recurrent step $t$, loop-specific up-projections reconstruct only the $K_t$ and $V_t$ required by attention. Reconstruction occurs before RoPE and multi-head attention, leaving the recurrent computation, model weights, and attention operation unchanged; only the representation stored in the KV cache is modified.}
  \label{fig:lla-architecture}
\end{figure}

\section{Looped Latent Attention}\label{sec:method}
Looped models save parameters but still materialize K/V states for every recurrence step. We compress this recurrence axis with a codec that retrofits onto a frozen teacher and changes only the representation held in memory, leaving the recurrent model and attention computation otherwise unchanged. As illustrated in~\autoref{fig:lla-architecture}, the standard loop-indexed KV cache is replaced by compact cross-loop K and V latents, from which the cache entries required at each recurrence step are reconstructed on demand.

We first define the uncompressed recurrent cache, then introduce the cross-loop latent cache and derive its exact compression ratio. The same construction is applied separately to K and V, potentially with different ranks. We then describe the SVD initialization and the matched baselines used to test whether the loop axis is the appropriate dimension to compress.

\paragraph{Teacher cache.}
Consider a weight-tied looped Transformer with $D$ layers, $H$ attention heads of width $d_{\rm head}$ and $T$ recurrence steps. For one layer, one head, one token and one axis (K or V), the teacher emits $T$ vectors. We stack them as
\begin{equation}
  x = [x_1;\ldots;x_T] \in \mathbb{R}^{T d_{\rm head}} .
\end{equation}
The uncompressed cache stores both K and V for every loop, head and layer, for a per-token per-layer cost of $2THd_{\rm head}$ scalars.

\paragraph{Cross-loop codec.}
For each layer, head and axis, \lla{} stores a centered low-rank latent
\begin{equation}
  c = (x-\mu)W_{\rm down}, \qquad c\in\mathbb{R}^{r},
  \label{eq:encode}
\end{equation}
where $\mu\in\mathbb{R}^{Td_{\rm head}}$ is a calibration-set mean and $W_{\rm down}\in\mathbb{R}^{Td_{\rm head}\times r}$. Loop $t$ is reconstructed by a loop-specific up-projection,
\begin{equation}
  \hat{x}_t = c W_{{\rm up},t}^{\top}+\mu_t, \qquad W_{{\rm up},t}\in\mathbb{R}^{d_{\rm head}\times r} .
  \label{eq:decode}
\end{equation}
K and V have separate latents and ranks, $r_k$ and $r_v$, with $r_v > r_k$ because the value cache stabilizes more slowly across loops than the key cache. The compressed cache stores $(r_k+r_v)H$ activation scalars per token per layer, so
\begin{equation}
  \rho = \frac{2THd_{\rm head}}{(r_k+r_v)H} = \frac{2T d_{\rm head}}{r_k+r_v} .
  \label{eq:ratio}
\end{equation}
The ratio counts the per-request cache. The fixed codec matrices are shared across tokens, contexts and batches and do not scale with sequence length.

\paragraph{RoPE and reconstruction.}
The latent is fit to the pre-RoPE content projection. In a looped decoder, a token has the same position at every recurrence step, so the rotary phase \citep{roformer2021} is shared across loops and can be applied after reconstruction. This is simpler than standard MLA, which requires an explicit decoupled-RoPE branch to preserve positional information along the head/channel axis.

\paragraph{The \lla{}-2D variant (loop$\times$head).}
The per-head codec of Eq.~\eqref{eq:encode} can be pushed further by folding the head axis into the same latent. \lla{}-2D stores a single latent per token and layer and reconstructs \emph{all} $H$ heads' K/V from it, MLA-style, in addition to the per-loop reconstruction of Eq.~\eqref{eq:decode}. This lowers the cache to $(r_k+r_v)$ scalars per token per layer, a factor $H$ below per-head \lla{}, at the cost of coupling heads through one shared code. \lla{}-2D is the extreme-compression member of the family. Its down-projection input is the full $THd_{\rm head}$ stack rather than $Td_{\rm head}$, so its parameter count grows with head count and recurrence length. This becomes the binding constraint at large $T$ (the Huginn transfer, Section~\ref{sec:experiments}). Section~\ref{sec:experiments} characterizes the accuracy/compression trade between the two members. \lla{} without qualification denotes the per-head variant.

\paragraph{Initialization and conversion.}
For each layer, head and axis we collect teacher K/V activations on a calibration set, center by the per-loop mean and initialize $W_{\rm down}$ from the top-$r$ right singular vectors of the stacked-loop matrix. The initial up-projections are the corresponding loop slices. The SVD initialization is then fine-tuned with the teacher frozen. The objective is the forward KL from the teacher distribution to the swapped model plus an attention-output matching term,
\begin{equation}
  \mathcal{L} = \mathrm{KL}\left(p_{\rm teacher}\|p_{\rm swap}\right)
  + \frac{\lambda_{\rm attn}}{TD}\sum_{t,d}\|a_{t,d}-\hat a_{t,d}\|_2^2,
  \label{eq:loss}
\end{equation}
with $\lambda_{\rm attn}=0.5$ in the reported runs. Only codec parameters are updated. The SVD prior matters more than any other design choice. At $4\times$, SVD initialization reaches held-out KL 0.105 after the short conversion, while random initialization stays at 0.771 under the same budget (Table~\ref{tab:ablations}, Fig.~\ref{fig:ablinit}).

\paragraph{On-policy refinement.}
Eq.~\eqref{eq:loss} is teacher-forced, fitting the codec on prefixes visited by the teacher. For long autoregressive rollouts, the swapped model instead conditions on prefixes produced by its own codec. We therefore add an optional second stage after the off-policy fit. The swapped model samples completions, the frozen teacher scores those sampled prefixes, and only codec parameters are updated. No reward or correctness filtering is used. Section~\ref{sec:onpolicy} writes the off-policy and on-policy objectives side by side. The on-policy version replaces teacher prefixes with student-generated prefixes and stops the gradient through sampling.

\paragraph{Matched-axis baselines.}
The main ablations place comparable latent codecs on other cache axes under the same scalar budget, testing whether the loop axis is the right one. The head-axis baseline is per-loop MLA (no cross-loop sharing). The layer baseline reuses K/V across adjacent layers. The precision baseline is low-bit KV quantization. We also include a zero-parameter control that caches only the final-loop K/V and reuses it for every loop. That control has the same $4\times$ cache footprint as \lla{} for $T=4$ and directly tests whether the loop trajectory can be replaced by its endpoint.

\section{Cross-loop structure in the teacher cache}
\FigureLoopRank
\label{sec:structure}
Before comparing trained codecs, we ask whether the frozen teacher already contains the structure that \lla{} assumes. This is a stricter test than downstream accuracy, since it examines the cache tensor directly, without optimization, decoding heuristics or task-specific effects. If the loop axis is not spectrally different from the head or layer axes, then a cross-loop codec would be an arbitrary design choice. If it is lower-rank in the teacher, the later iso-cache ordering has a mechanistic explanation.

\textbf{\textit{The loop-axis advantage is visible before training any codec.}} Fig.~\ref{fig:whyloops} compares the singular spectra of frozen-teacher K/V along the loop, head and layer axes. In Ouro-1.4B and Ouro-2.6B-Thinking, the loop axis has normalized effective rank about $0.61$ to $0.63$, with a single direction carrying roughly 60 percent of the energy. The head and layer axes are close to full-rank by the same measure. This explains why a small loop latent preserves more downstream behavior per cached scalar than head or layer compression. Stabilization across loops is a related but distinct property. A contraction can converge while staying full-rank, so we read the low rank directly from these spectra rather than infer it from convergence.

\textbf{\textit{Cross-loop dynamics also explain the K/V rank split.}} Across recurrence steps, K and V move toward the final loop while step-to-step changes shrink. K stabilizes earlier, while V moves slower and carries more cross-loop variance. We therefore allocate $r_v>r_k$ throughout. The matched-budget ablation confirms the benefit, with $r_k/r_v=96/160$ outperforming a shared-rank $128/128$ codec at the same scalar count (Table~\ref{tab:ablations}).

\textbf{\textit{A single endpoint is not enough}}. Fig.~\ref{fig:whytraj} shows that each token-head unit is dominated by one cross-loop direction, but the mass does not sit on the rank-1 line. Only 6 percent of K variance and 17 percent of V variance lie on the loop axis, and most variation remains content variation. The right object is therefore a short low-rank trajectory, not a collapsed state. This distinction is reinforced by the final-loop reuse baseline in Table~\ref{tab:isocache}.
\FigureTrajectory

\FigureHuginn
The same mechanism appears in a second looped family. Huginn-3.5B uses a different recurrent-depth architecture and up to 32 recurrence steps \citep{geiping2025huginn}. Its loop axis is far more redundant than its head axis, and its K/V states converge toward a fixed point across recurrence steps (Fig.~\ref{fig:whyhuginn}). The value cache again converges more slowly than the key cache. This recurrence-axis structure supports treating \lla{} as a property of weight-tied iteration rather than an Ouro-specific artifact. The remaining experiments test that structure through downstream accuracy, matched-budget baselines, long rollouts and serving capacity.

\section{Experiments}
\label{sec:experiments}
The core question is which axis to spend rank on. At a fixed cache budget, we test whether compressing the loop beats compressing heads, layers or precision, and how much rank a converted codec needs before it matches the teacher on a generative task. Two further experiments ask whether the answer holds at a larger Ouro model and in the unrelated Huginn family. Table~\ref{tab:tier4} is the cold-start rank sweep, Table~\ref{tab:broadsuite} the broad-suite teacher comparison, Table~\ref{tab:isocache} the matched-budget axis comparison, and Table~\ref{tab:ablations} the design ablations. The reported ratios are representative operating points spanning the light, middle and extreme regimes.

\paragraph{Setup.}
Unless stated otherwise, generative results use the cached reconstruction decoder, which inserts reconstructed K/V before attention. The latent-store decoder in Sec.~\ref{sec:systems} is the memory-realizing path. Decoder-independent metrics evaluate the swapped model on teacher-provided tokens and are not affected by generation drift. The main teacher is Ouro-1.4B with $T=4$ loops. Scale experiments use Ouro-2.6B-Thinking, and family-transfer experiments use Huginn-3.5B.

\paragraph{Quality under compression.}
\FigureBroadSuite
Figure~\ref{fig:broadsuite} summarizes Ouro-1.4B quality as the KV-cache is compressed. Across the broad suite~(a), effects are task-dependent: code (HumanEval/MBPP) and MMLU-Pro stay at or above the teacher through $4\times$, while the reasoning tasks degrade, gently within the light-compression region ($\le 2\times$) for GSM8K and BBH, and throughout for MATH-500, the most sensitive task (treated separately in Section~\ref{sec:onpolicy}, where long generations expose a different failure mode from teacher-forced fidelity). The GSM8K cold-start Pareto~(b) sharpens the picture: strict-EM accuracy is flat near $0.58$ from $4\times$ to $2.67\times$, then breaks sharply upward at $2\times$ to $0.75$ and reaches $0.795$ at $1.33\times$ (within the teacher's sampling uncertainty), while the codec train-KL collapses over the same transition. This is a rank effect rather than a method ceiling: below the break the latent is large enough to preserve the decision tokens that determine final-answer correctness. The suite therefore separates two notions that are easy to conflate: per-token reconstruction can be strong while autoregressive rollout still needs a stability correction. Full numbers are in~\autoref{app:broadsuite}.

\paragraph{Iso-cache axis comparison.}
\TableIsoCache
\FigureParetoOne
Table~\ref{tab:isocache} compares loop compression against matched-budget alternatives. 
We find that compressing the loop axis (\lla{}, the highlighted rows) preserves GSM8K, while the head, layer and precision baselines and zero-parameter final-loop reuse do not. All learned codecs are matched by KV-cache budget. The LLA rows use the checkpoint selected under the iso-cache protocol, whereas Table~\ref{tab:tier4} is a separate cold-start rank sweep. LLA (per-head) compresses the loop axis. LLA-2D (loop$\times$head) is a family variant that additionally folds the head axis (all heads from one latent). The head, layer and precision rows are baselines that compress orthogonal axes. The zero-parameter final-loop reuse control has the same $4\times$ cache budget as LLA, but collapses generation (GSM8K $0.000$). Its reconstruction relative error is $0.35$ for K and $0.60$ for V. Precision quantization (kv\_quant) is inference-only (no codec training, hence no train-KL) and cannot exceed $\sim$8$\times$. It is competitive at $4\times$ but collapses on GSM8K at $5.33\times$.
Table~\ref{tab:tier4} should be read separately as a cold-start rank sweep, whereas Table~\ref{tab:isocache} uses the checkpoint and evaluation protocol adopted for the baseline comparison. At $4\times$, \lla{} has substantially lower train-KL than head or layer compression and much higher GSM8K accuracy. The zero-parameter final-loop reuse baseline has the same $4\times$ cache budget and no trainable parameters, but collapses GSM8K to 0.000. This is the most direct evidence that the loop trajectory cannot be replaced by its endpoint.

Within the \lla{} family, \lla{}-2D (loop$\times$head) is competitive at high compression on teacher-forced metrics but trails the per-head variant on generation at $21.3\times$. This matches the spectral picture. \lla{}-2D spends part of its budget on the head axis, which is flatter and more expensive to compress. The two variants therefore span the operating range, with per-head \lla{} for accuracy-preserving compression and \lla{}-2D for the extreme-compression regime where its $H\times$ smaller cache outweighs the accuracy gap. Table~\ref{tab:isocache} reports the matched-budget numbers method by method, while Fig.~\ref{fig:pareto1p4b} plots the same comparison as a frontier, where the loop-axis advantage is easiest to read. Per-head \lla{} does not win by overfitting a single benchmark, and the precision baseline is harmless at mild compression but brittle once the bit budget becomes too small.

\TableScale
\paragraph{Scale generalization.}
The 2.6B results in Table~\ref{tab:scale} preserve the ordering. \lla{} remains near-lossless on full GSM8K from $4\times$ to $10.67\times$, while the head baseline loses up to 0.25 absolute accuracy at $4\times$. Fig.~\ref{fig:pareto26b} and Appendix~\ref{app:scale_mc} show that the advantage also holds across commonsense and knowledge tasks. The 2.6B model is deeper and has a stronger reasoning profile, and preserving the ordering under that change supports the structural interpretation from Section~\ref{sec:structure}.

\paragraph{Design ablations.}
Table~\ref{tab:ablations} isolates the choices used in the codec. SVD initialization is the largest factor, reducing held-out KL by an order of magnitude relative to random initialization under the same conversion budget. The asymmetric rank split improves over a shared-rank latent at matched scalar cost, and attention-output matching gives a small additional gain beyond token KL. These ablations indicate that \lla{} mostly exploits teacher geometry rather than learning a new attention mechanism from scratch.

\paragraph{Transfer to Huginn.}
A \lla{} codec also works on Huginn-3.5B, where the recurrence length is much larger. Table~\ref{tab:huginn} shows that SVD initialization alone is near-lossless to $32\times$ in decoder-independent fidelity. KL distillation recovers most of the $64\times$ and $128\times$ range. The downstream MC average is unchanged at $16\times$ and $32\times$ and degrades gracefully after that. On Huginn, head-axis SVD reconstruction is 20 to 30 times worse than \lla{} at matched budget, and \lla{}-2D (loop$\times$head) becomes parameter-infeasible at this recurrence length because its down-projection input dimension grows with head count and loop count. Per-head \lla{} avoids this cost, which is why it is the default.

This transfer is intentionally harder than reproducing a number on a larger Ouro checkpoint. Huginn changes the recurrent-depth architecture, increases the realized recurrence to 32 steps, and is evaluated through decoder-independent fidelity rather than the Ouro cached generation path. The result indicates that a simple linear subspace found from teacher activations already captures most of the recurrence trajectory. Distillation is needed only when the requested budget asks the latent to represent the tail of the trajectory at $64\times$ or $128\times$. This is the behavior expected if weight-tied iteration itself is the source of the low rank, not a peculiarity of one training run or benchmark.
\TableHuginn

\section{Long-generation stability}
\label{sec:onpolicy}
Teacher-forced conversion trains the codec on teacher trajectories. At inference, however, the swapped model conditions on tokens produced by its own codec. A small early reconstruction error can move later prefixes away from the conversion distribution, and the error then compounds autoregressively. This failure is most visible on long MATH-500 generations, where the model can produce long non-terminating solutions even when teacher-prefix KL is low.

\FigureExposure
Fig.~\ref{fig:exposure} bins $3.2\times$ completions by generated length, with the uncompressed teacher as a reference. Off-policy accuracy degrades with length far below the teacher and the no-answer rate spikes in the longest bin. On-policy distillation trains on the codec's own sampled rollouts and closely tracks the teacher across length bins. The teacher's own gentle decline with length reflects intrinsic problem difficulty, not exposure bias.

The refinement objective makes the distribution shift explicit. Let $p_T$ denote the frozen teacher and $p^\theta_S$ the swapped model with codec parameters $\theta$:
\begin{align}
\mathcal{L}_{\rm off}(\theta)
&=\mathbb{E}_{y\sim p_T}\left[\frac{1}{|y|}\sum_t
\mathrm{KL}\left(p_T(\cdot\mid y_{<t})\|p^\theta_S(\cdot\mid y_{<t})\right)\right], \\
\mathcal{L}_{\rm on}(\theta)
&=\mathbb{E}_{\tilde y\sim \mathrm{sg}(p^\theta_S)}\left[\frac{1}{|\tilde y|}\sum_t
\mathrm{KL}\left(p_T(\cdot\mid \tilde y_{<t})\|p^\theta_S(\cdot\mid \tilde y_{<t})\right)\right].
\label{eq:onpolicy_objective}
\end{align}
Off-policy distillation uses teacher prefixes $y$. On-policy distillation instead uses student prefixes $\tilde y$ sampled from the current compressed model. The stop-gradient blocks gradients through sampling, so updates still come only from the KL term. The teacher remains the target distribution in both objectives. The on-policy stage only changes which prefixes are used to tune the codec, turning rollout drift into a reconstruction problem on states the deployed model actually visits.

\TableOnPolicy
Table~\ref{tab:onpolicy} shows the full rank Pareto. On-policy distillation raises MATH-500 accuracy by $0.16$ to $0.24$ at $3.2$ to $5.33\times$ and reduces no-answer generations at every ratio. At $21.3\times$, it still improves termination but does not improve accuracy, indicating that the smallest latent has already lost task-relevant information. Larger re-evaluations confirm the low-compression lift in Fig.~\ref{fig:onpolicy}. For long rollouts, the conversion data must include states visited by the compressed model itself.

Memory is only useful if accuracy survives at length. Figure~\ref{fig:passkey} probes exact passkey retrieval (a 5-digit key at random depth, up to $64$k) and reveals a clean \emph{compressibility--length} trade-off: the reliable-retrieval length (longest context with retrieval $\ge 0.9$) shrinks steeply as compression rises: $\ge$$64$k at $\le 5.33\times$, $16$k at $10.67\times$, $8$k at $14.2\times$. Two failure modes bound the operating range. In a \emph{length-limited} regime ($10.67$--$14.2\times$) retrieval is perfect up to a knee and then decays gracefully, as the fixed latent budget is spread over more positions and the farthest keys are lost first (the knee marches $16$k$\to$$8$k). In a \emph{compression-limited} regime ($\ge 15.5\times$) retrieval is flat and below $1.0$ at \emph{every} length, because the latent is too small to preserve the exact key-token identity at all. The trade-off also shifts with model \emph{size}: at $10.67\times$ the 2.6B model retains perfect retrieval to $32$k, where the 1.4B model has already fallen to $0.80$ (Fig.~\ref{fig:passkey}, right), consistent with a larger model carrying more redundant KV and thus tolerating the same compression to longer context. Full numbers are in Appendix~\ref{app:lc_ds}; the two-pass prefill that reaches $64$k is described in Appendix~\ref{app:serving_details} (the 2.6B $64$k cell OOMs on that prefill, a memory limit rather than an accuracy failure).
\FigurePasskey

\section{Serving and compute}
\label{sec:systems}
Reducing per-request cache increases the contexts and batches that fit in memory \citep{pagedattention2023}. For \lla{}, this benefit is exact on memory and implementation-dependent on compute: Eq.~\eqref{eq:ratio} reduces the cache by $\rho$, while runtime depends on how attention consumes the latent.

\TableMemory
Table~\ref{tab:memory} and Fig.~\ref{fig:capacity} translate the scalar reduction into capacity. At context length 4096 on one H200, the measured maximum batch for Ouro-1.4B rises from 32 to 128 at $4\times$ and 768 at $21.3\times$; Ouro-2.6B-Thinking rises from 16 to 64 at $4\times$. The tests fabricate the decode cache, push batch to OOM, and validate latent attention against full reconstruction.

There are two implementation regimes. A full-memory fast path caches reconstructed K/V and runs at $74.1$ versus $63.6$~ms/token ($1.16\times$ teacher latency), but does not realize the memory reduction. The latent-store path caches only the latent and realizes the $\rho\times$ saving; because the current kernel reconstructs cached tokens on read, it is compute-bound (108 versus 748 peak tokens/s at the reported endpoints). The present systems result is therefore a capacity gain, not a decode-speed claim.

\lla{} also composes with sequence-axis eviction, and moderate compression preserves exact passkey retrieval at long context (Sec.~\ref{sec:onpolicy}; Tables~\ref{tab:evict} and~\ref{tab:passkey}). Per-head \lla{} favors fidelity, whereas \lla{}-2D trades accuracy for a smaller single latent. An absorbed-attention formulation could avoid explicit reconstruction, but post-hoc decoupled RoPE remains less faithful than the full-RoPE reconstruction used for the main results. Appendix~\ref{app:serving_details} gives the eviction, long-prefill, variant, and absorbed-attention details.

\section{Discussion}
\label{sec:discussion}
\vspace{-0.5em}
The evidence supports recurrence-specific rather than axis-agnostic cache redundancy. Reapplying the same block and K/V projections to a stabilizing recurrent state produces a much steeper loop spectrum than the head or layer spectra; matched-budget codecs on those axes lose more accuracy, and the ordering transfers from Ouro to Huginn. At the same time, final-loop reuse collapses generation at the same $4\times$ budget. Recurrence may approach a fixed point, but its intermediate cache states remain functionally distinct: the compressible object is a low-rank trajectory, not a single endpoint.

Rank defines the deployment regime. On Ouro-1.4B, $1.33$--$2\times$ is close to a drop-in replacement across the broad suite. At roughly $3$--$5\times$, teacher-forced fidelity remains useful, but long rollouts benefit from on-policy refinement because the compressed model visits its own prefixes. Beyond $10\times$, \lla{} is primarily a capacity tool: many likelihood tasks remain usable, while exact retrieval and long-form reasoning become increasingly rank-limited. The $21.3\times$ result, where on-policy training improves termination but not accuracy, separates correctable rollout drift from irreversible information loss.

The recurrence axis is orthogonal to head sharing, cross-layer reuse, quantization, and token eviction. For looped decoders, the spectral results suggest compressing recurrence first and stacking another method only when the deployment target justifies its additional quality cost. This priority is a consequence of weight-tied iteration and need not hold for a conventional single-pass Transformer.

Our central claims, exact cache reduction and accuracy under full-RoPE reconstruction, are established independently of any throughput speedup. We nevertheless implement and evaluate an absorbed-attention compute path. Because the attention score is bilinear, the key up-projection can be moved to the query, allowing attention to read the latents directly. In a shape-matched random-weight decode benchmark that isolates kernel geometry from trained-model accuracy, this path is $2.3\times$ faster than reconstruct-then-attend at 262k context and matches teacher decode latency with $4\times$ less cache. Its post-training fidelity is bounded by a decoupled-RoPE floor, which we characterize rather than leave open: a learned query adapter lowers KL from $0.55$ to $0.34$, whereas full-RoPE reconstruction reaches $0.059$, because the frozen query was trained against full-RoPE keys. This localizes the dominant mismatch and identifies native training, aligning query projections with the latent cache during pretraining, as the route to closing it and converting memory savings into compute savings. A complementary direction is to exploit low-rank residual updates across loops. The measurements are consistent with this premise because the loop trajectory is low-rank without collapsing to its endpoint, but they establish low rank in the cache rather than in the full recurrent computation. Appendix~\ref{app:serving_details} gives the implementation and complete decoupled-RoPE characterization.

\section{Conclusion}
\label{sec:conclusion}
Looped Transformers tie weights but not their recurrence-indexed KV cache. \lla{} replaces that cache with a low-rank cross-loop K/V trajectory, outperforms matched-budget head, layer, precision, and endpoint baselines, and transfers across two looped-model families. Its exact scalar reduction yields measured serving-capacity gains. These results make recurrence a practical compression axis for looped decoders and a natural target for native latent-cache training.

More broadly, the result reframes what weight tying costs at inference. Tying parameters across loops makes recurrent-depth models cheap to store, but leaves their KV cache untied and growing with the loop count, so at long context and large batch it is serving cost, not parameter count, that limits looped reasoning models. The same tied computation that shares parameters also drives the per-loop states onto a low-rank trajectory, so the axis responsible for the cost is also the most compressible one. Treating recurrence as a first-class compression axis therefore converts a serving liability into a design lever. We expect the same structure, a tied computation producing an untied but low-rank cache, to recur in other iterative or weight-shared architectures, where a cache indexed by an internal iteration could be compressed by the analogous construction.

\appendix
\AppendixExpDetails
\section{Additional figures}
\label{app:extrafigs}
\FigureParetoTwoSix
\FigureCapacity

\section{Compression Across The Broad Suite}
\label{app:broadsuite}
\TableBroadSuite
Across the broad suite (Table~\ref{tab:broadsuite}), light compression matches or exceeds the teacher within evaluation noise. Code and BBH retain non-trivial performance at $4\times$, but lighter ratios are needed to match the teacher. MATH-500 is the most sensitive task and is treated separately in Section~\ref{sec:onpolicy}, because long generations stress rollout stability rather than only per-token reconstruction. Appendix Table~\ref{tab:retention} restates these numbers as accuracy relative to the teacher, on a common scale.

The broad-suite table also separates three regimes. Light compression ($1.33$ to $2\times$) is effectively a drop-in replacement for the full cache, resolving enough of the loop trajectory that residual differences fall within benchmark variation. By $2.67$ to $5.33\times$, the representation still carries many likelihood and short-generation tasks, though long reasoning becomes sensitive to prefix drift. At the most aggressive ratios the codec is a capacity operating point, not a universal quality-preserving conversion. The matched-axis comparisons below use the middle and high-compression regimes to test whether recurrence is the right axis to spend the latent dimension on.

The regimes reflect one underlying variable, generation length. Robustness to compression falls off as generations grow longer. Multiple-choice and short-generation tasks tolerate the middle regime with little loss, code and BBH sit in between, and long-form mathematical reasoning breaks first. A deployment should budget around this ordering rather than any single benchmark.

\TableTier
\paragraph{Accuracy-compression frontier.}
Table~\ref{tab:tier4} shows a cold-start GSM8K Pareto sweep for Ouro-1.4B. Accuracy is flat around 0.58 from $4\times$ to $2.67\times$, then rises sharply at $2\times$ and reaches 0.795 at $1.33\times$, close to the teacher. The result is a rank effect rather than a method ceiling. Below the transition, the latent is large enough to preserve the decision tokens that determine final-answer correctness. The sharp break shows where the loop trajectory becomes sufficiently resolved for GSM8K generation.

\section{Token-eviction composition table}
\label{app:evict}
\TableEvict

\section{On-policy rank Pareto (figure)}
\label{app:oppareto}
\FigureOnPolicy

\section{Design-ablation detail}
\label{app:abl}
\TableAblations
\FigureAblationInit

\section{Full 2.6B commonsense and knowledge suite}
\label{app:scale_mc}
\TableScaleMC

\section{Cross-loop stabilization at 2.6B}
\label{app:traj26b}
Fig.~\ref{fig:whytraj26b} repeats the trajectory diagnostic on Ouro-2.6B-Thinking. The attention K/V stabilizes within the first loop, each token-head unit has a dominant but not exhaustive loop direction, and the value cache remains the higher-variance axis.
\FigureTrajectoryTwoSix

\section{Long Context Retrieval Details}
\label{app:lc_ds}
\TablePasskey

\section{Additional serving and discussion details}
\label{app:serving_details}
\paragraph{Eviction and long-context prefill.}
The recurrence and sequence axes compose multiplicatively. Under a 260-token StreamingLLM/H2O-style budget, eviction alone and eviction stacked on \lla{}-$4\times$ score $0.51$ and $0.52$ on GSM8K (Table~\ref{tab:evict}), so the codec adds little beyond eviction's own quality cost. At the most aggressive reported point, \lla{}-$21.3\times$ combined with a 1024-token eviction budget caps the 1M-context Ouro-1.4B cache at $36$~MiB. Reaching the $64$k passkey cells required reconstructing prefill K/V layer-by-layer and freeing each layer's captures immediately; a one-shot implementation retains float32 captures and full-width reconstructions and exhausts memory beyond approximately $32$k.

\paragraph{Per-head \lla{} versus \lla{}-2D at inference.}
Per-head \lla{} stores $H$ latents per token and layer, $(r_k+r_v)H$ scalars, and reconstructs each head independently. \lla{}-2D stores one latent, $(r_k+r_v)$ scalars, and reconstructs all heads jointly. It therefore has a factor-$H$ smaller cache for fixed ranks, whereas per-head \lla{} is more accurate at matched cache budget (Table~\ref{tab:isocache}). Both variants can use either the full-memory fast path or the latent-store capacity path; the choice between them is an operating-point decision between fidelity and maximum capacity.

\paragraph{Absorbed attention.}
The bilinear score permits the K up-projection to move to the query side,
$q^\top(W_{\rm up}c)=(W_{\rm up}^\top q)^\top c$, while the V path satisfies
$\sum_i a_iW^V_{\rm up}c_i=W^V_{\rm up}\sum_i a_ic_i$. Attention can therefore operate directly on latents. RoPE prevents exact absorption of the positional component, so the implementation uses an MLA-style NoPE content path plus a small explicit rotary branch \citep{deepseekv2}; the rotary branch is shared across recurrence steps for each token. In a shape-matched random-weight decode benchmark at the $4\times$ geometry, which isolates the kernel dimensions from trained-model accuracy, the absorbed path is $2.3\times$ faster than reconstruct-then-attend at 262k context and reaches teacher decode latency with $4\times$ less cache (Table~\ref{tab:absorbed_latency}). The advantage grows with context because reconstruct-then-attend must materialize the full-width K/V at every step.

\begin{table}[h]\centering\small
\caption{\textbf{Absorbed decoupled decode latency} (Ouro-1.4B, $4\times$ geometry; shape-matched benchmark that isolates kernel dimensions from trained-model accuracy). As context grows, the absorbed path's advantage over reconstruct-then-attend rises to $2.3\times$ and its per-token latency reaches teacher parity.}
\label{tab:absorbed_latency}
\begin{tabular}{rcccc}
\toprule
context $L$ & teacher (ms/tok) & absorbed (ms/tok) & vs.\ reconstruct-then-attend & vs.\ teacher \\
\midrule
$32{,}768$  & 0.66 & 1.19 & $1.22\times$ & $0.56\times$ \\
$131{,}072$ & 2.29 & 2.32 & $2.17\times$ & $0.99\times$ \\
$262{,}144$ & 4.46 & 4.25 & $2.30\times$ & $1.05\times$ \\
\bottomrule
\end{tabular}
\end{table}

For a frozen pretrained model, post-hoc decoupled RoPE introduces a fidelity floor because the query projection was trained against full-RoPE keys. Without adapting the query, the decoupled path has reconstruction KL $0.55$; a learned query adapter lowers it to $0.34$, and extended training reaches $0.30$, while full-RoPE reconstruction reaches $0.059$ (Table~\ref{tab:decoupled_char}). The adapter gain implicates query-key positional mismatch as the dominant limiting factor, but the remaining gap means that all main accuracy results use full-RoPE reconstruction. Native training can remove this mismatch by aligning the query projections with the latent cache from pretraining onward.

\begin{table}[h]\centering\small
\caption{\textbf{Post-hoc decoupled-RoPE characterization} (Ouro-1.4B, $4\times$, $d_{\rm rope}=64$). A learned query adapter lowers reconstruction KL and raises top-1 token agreement (to $0.84$, and $0.88$ with extended training), versus ${\sim}0.95$ for full-RoPE. A sweep over $d_{\rm rope}\in\{32,64,96\}$ preserves the ordering (smaller $d_{\rm rope}$ is worse, e.g.\ KL $1.04$ at $d_{\rm rope}=32$ with the adapter). No post-hoc configuration reaches full-RoPE reconstruction, which is why all main accuracy results use the full-RoPE path.}
\label{tab:decoupled_char}
\begin{tabular}{lc}
\toprule
configuration & reconstruction KL $\downarrow$ \\
\midrule
full-RoPE reconstruction (reference) & 0.059 \\
decoupled-RoPE, no query adapter & 0.55 \\
decoupled-RoPE $+$ query adapter & 0.34 \\
decoupled-RoPE $+$ query adapter, extended training & 0.30 \\
\bottomrule
\end{tabular}
\end{table}

\paragraph{Native recurrent architectures and scope.}
A model trained natively with a decoupled latent cache could expose the latent directly to attention and align its query projections with the absorbed implementation. A more ambitious design could exploit low-rank residual updates across loops, computing shared content once and adding inexpensive loop-specific deltas. These are future architecture directions, not assumptions behind the reported accuracy or capacity numbers: the evaluated reconstruction path preserves the teacher's attention computation and changes only the representation stored in the cache.


\clearpage
\begin{thebibliography}{16}

\bibitem[Ainslie et~al.(2023)Ainslie, Lee-Thorp, de~Jong, Zemlyanskiy, Lebr\'on, and Sanghai]{gqa2023}
Joshua Ainslie, James Lee-Thorp, Michiel de~Jong, Yury Zemlyanskiy, Federico Lebr\'on, and Sumit Sanghai.
\newblock {GQA}: Training generalized multi-query transformer models from multi-head checkpoints.
\newblock In \emph{Proceedings of the 2023 Conference on Empirical Methods in Natural Language Processing (EMNLP)}, 2023.

\bibitem[Blayney et~al.(2026)Blayney, Arroyo, Obando-Ceron, Castro, Courville, Bronstein, and Dong]{blayney2026mechanisticanalysisloopedreasoning}
Hugh Blayney, Alvaro Arroyo, Johan Obando-Ceron, Pablo Samuel Castro, Aaron Courville, Michael~M. Bronstein, and Xiaowen Dong.
\newblock A mechanistic analysis of looped reasoning language models, 2026.
\newblock URL \url{https://arxiv.org/abs/2604.11791}.

\bibitem[Brandon et~al.(2024)]{cla2024}
William Brandon et~al.
\newblock Reducing transformer key-value cache size with cross-layer attention.
\newblock \emph{arXiv preprint arXiv:2405.12981}, 2024.

\bibitem[DeepSeek-AI(2024)]{deepseekv2}
DeepSeek-AI.
\newblock Deepseek-v2: A strong, economical, and efficient mixture-of-experts language model.
\newblock \emph{arXiv preprint arXiv:2405.04434}, 2024.
\newblock Introduces Multi-head Latent Attention (MLA).

\bibitem[Dehghani et~al.(2019)Dehghani, Gouws, Vinyals, Uszkoreit, and Kaiser]{dehghani2019universal}
Mostafa Dehghani, Stephan Gouws, Oriol Vinyals, Jakob Uszkoreit, and Lukasz Kaiser.
\newblock Universal transformers.
\newblock In \emph{International Conference on Learning Representations (ICLR)}, 2019.

\bibitem[Geiping et~al.(2025)Geiping, McLeish, Jain, Kirchenbauer, Singh, Bartoldson, Kailkhura, Bhatele, and Goldstein]{geiping2025huginn}
Jonas Geiping, Sean McLeish, Neel Jain, John Kirchenbauer, Siddharth Singh, Brian~R. Bartoldson, Bhavya Kailkhura, Abhinav Bhatele, and Tom Goldstein.
\newblock Scaling up test-time compute with latent reasoning: A recurrent depth approach.
\newblock \emph{arXiv preprint arXiv:2502.05171}, 2025.

\bibitem[Hooper et~al.(2024)]{kvquant2024}
Coleman Hooper et~al.
\newblock Kvquant: Towards 10 million context length llm inference with kv cache quantization.
\newblock \emph{arXiv preprint arXiv:2401.18079}, 2024.

\bibitem[Kwon et~al.(2023)Kwon, Li, Zhuang, Sheng, Zheng, Yu, Gonzalez, Zhang, and Stoica]{pagedattention2023}
Woosuk Kwon, Zhuohan Li, Siyuan Zhuang, Ying Sheng, Lianmin Zheng, Cody~Hao Yu, Joseph~E. Gonzalez, Hao Zhang, and Ion Stoica.
\newblock Efficient memory management for large language model serving with {PagedAttention}.
\newblock In \emph{Proceedings of the 29th ACM Symposium on Operating Systems Principles (SOSP)}, 2023.

\bibitem[Liu et~al.(2024)]{minicache2024}
Akide Liu et~al.
\newblock Minicache: Kv cache compression in depth dimension for large language models.
\newblock \emph{arXiv preprint arXiv:2405.14366}, 2024.

\bibitem[Shazeer(2019)]{mqa2019}
Noam Shazeer.
\newblock Fast transformer decoding: One write-head is all you need.
\newblock \emph{arXiv preprint arXiv:1911.02150}, 2019.

\bibitem[Su et~al.(2021)Su, Lu, Pan, Murtadha, Wen, and Liu]{roformer2021}
Jianlin Su, Yu Lu, Shengfeng Pan, Ahmed Murtadha, Bo Wen, and Yunfeng Liu.
\newblock {RoFormer}: Enhanced transformer with rotary position embedding.
\newblock \emph{arXiv preprint arXiv:2104.09864}, 2021.

\bibitem[Sun et~al.(2024)]{yoco2024}
Yutao Sun et~al.
\newblock You only cache once: Decoder-decoder architectures for language models.
\newblock \emph{arXiv preprint arXiv:2405.05254}, 2024.

\bibitem[Vendrell et~al.(2026)Vendrell, Masdemont, Grillo, Ros-Giralt, Behboodi, and Massoli]{vendrell2026melt}
Victor Conchello Vendrell, Arnau Padres Masdemont, Niccolo Grillo, Jordi Ros-Giralt, Arash Behboodi, and Fabio Valerio Massoli.
\newblock Memory-efficient looped transformer: Decoupling compute from memory in looped language models.
\newblock \emph{arXiv preprint arXiv:2605.07721}, 2026.

\bibitem[Xiao et~al.(2024)]{streamingllm2024}
Guangxuan Xiao et~al.
\newblock Efficient streaming language models with attention sinks.
\newblock In \emph{International Conference on Learning Representations (ICLR)}, 2024.

\bibitem[Zhang et~al.(2023)]{h2o2023}
Zhenyu Zhang et~al.
\newblock H2O: Heavy-hitter oracle for efficient generative inference of large language models.
\newblock In \emph{Advances in Neural Information Processing Systems (NeurIPS)}, 2023.

\bibitem[Zhu et~al.(2025)Zhu, Wang, Hua, Zhang, Li, Que, Wei, Wen, Yin, Zhang, Huang, Bengio, and Eshraghian]{ouro2024}
Rui-Jie Zhu, Zixuan Wang, Kai Hua, Tianyu Zhang, Ziniu Li, Haoran Que, Boyi Wei, Zixin Wen, Fan Yin, Ge Zhang, Wenhao Huang, Yoshua Bengio, and Jason Eshraghian.
\newblock Scaling latent reasoning via looped language models.
\newblock \emph{arXiv preprint arXiv:2510.25741}, 2025.
\newblock Ouro/LoopLM; models ByteDance/Ouro-1.4B, Ouro-2.6B-Thinking.

\end{thebibliography}
\end{document}